\documentclass[a4paper,fleqn]{cas-sc}

\usepackage[]{natbib}
\newcommand{\tabincell}[2]{\begin{tabular}{@{}#1@{}}#2\end{tabular}}

\def\tsc#1{\csdef{#1}{\textsc{\lowercase{#1}}\xspace}}
\tsc{WGM}
\tsc{QE}
\tsc{EP}
\tsc{PMS}
\tsc{BEC}
\tsc{DE}

\newtheorem{defn}{Definition}[section]
\newtheorem{prob}{Problem}[section]
\begin{document}

\let\WriteBookmarks\relax
\def\floatpagepagefraction{1}
\def\textpagefraction{.001}
\shorttitle{Vessel Trajectory Similarity Computation}
\shortauthors{M. Liang et al.}

\title [mode = title]{An Unsupervised Learning Method with Convolutional Auto-Encoder for Vessel Trajectory Similarity Computation}


%
%
\author[1,2]{Maohan Liang}
\ead{mhliang@whut.edu.cn}
\author[1,2]{Ryan Wen Liu}
\cormark[2]
\ead{wenliu@whut.edu.cn}
%
%
\author[3]{Shichen Li}
\ead{271755@whut.edu.cn}
\author[4]{Zhe Xiao}
\ead{xiaoz@ihpc.a-star.edu.sg}
%
\author[5]{Xin Liu}
\ead{xin.liu@aist.go.jp}
%
%
%
\author[2]{Feng Lu}
\cormark[2]
\ead{luf@lreis.ac.cn}
%
%
\address[1]{Hubei Key Laboratory of Inland Shipping Technology, School of Navigation, Wuhan University of Technology, Wuhan 430063, China}
\address[2]{State Key Laboratory of Resources and Environmental Information System, Institute of Geographic Sciences and Natural Resources Research, CAS, Beijing 100101, China}
\address[3]{School of Computer Science and Technology, Wuhan University of Technology, Wuhan 430063, China}
\address[4]{Institute of High Performance Computing, A*Star (Agency for Science, Technology and Research), CO, 118411, Singapore}
\address[5]{AIRC/RWBC-OIL, AIST, Tokyo 135-0064, Japan}
%
\cortext[cor2]{Corresponding authors.}


\begin{abstract}
	To achieve reliable mining results for massive vessel trajectories, one of the most important challenges is how to efficiently compute the similarities between different vessel trajectories. The computation of vessel trajectory similarity has recently attracted increasing attention in the maritime data mining research community. However, traditional shape- and warping-based methods often suffer from several drawbacks such as high computational cost and sensitivity to unwanted artifacts and non-uniform sampling rates, etc. To eliminate these drawbacks, we propose an unsupervised learning method which automatically extracts low-dimensional features through a convolutional auto-encoder (CAE). In particular, we first generate the informative trajectory images by remapping the raw vessel trajectories into two-dimensional matrices while maintaining the spatio-temporal properties. Based on the massive vessel trajectories collected, the CAE can learn the low-dimensional representations of informative trajectory images in an unsupervised manner. The trajectory similarity is finally equivalent to efficiently computing the similarities between the learned low-dimensional features, which strongly correlate with the raw vessel trajectories. Comprehensive experiments on realistic data sets have demonstrated that the proposed method largely outperforms traditional trajectory similarity computation methods in terms of efficiency and effectiveness. The high-quality trajectory clustering performance could also be guaranteed according to the CAE-based trajectory similarity computation results.
\end{abstract}
%
%
%
\maketitle
%
%
%
\begin{figure}[t]	
	\centering
	\includegraphics[width=1\linewidth]{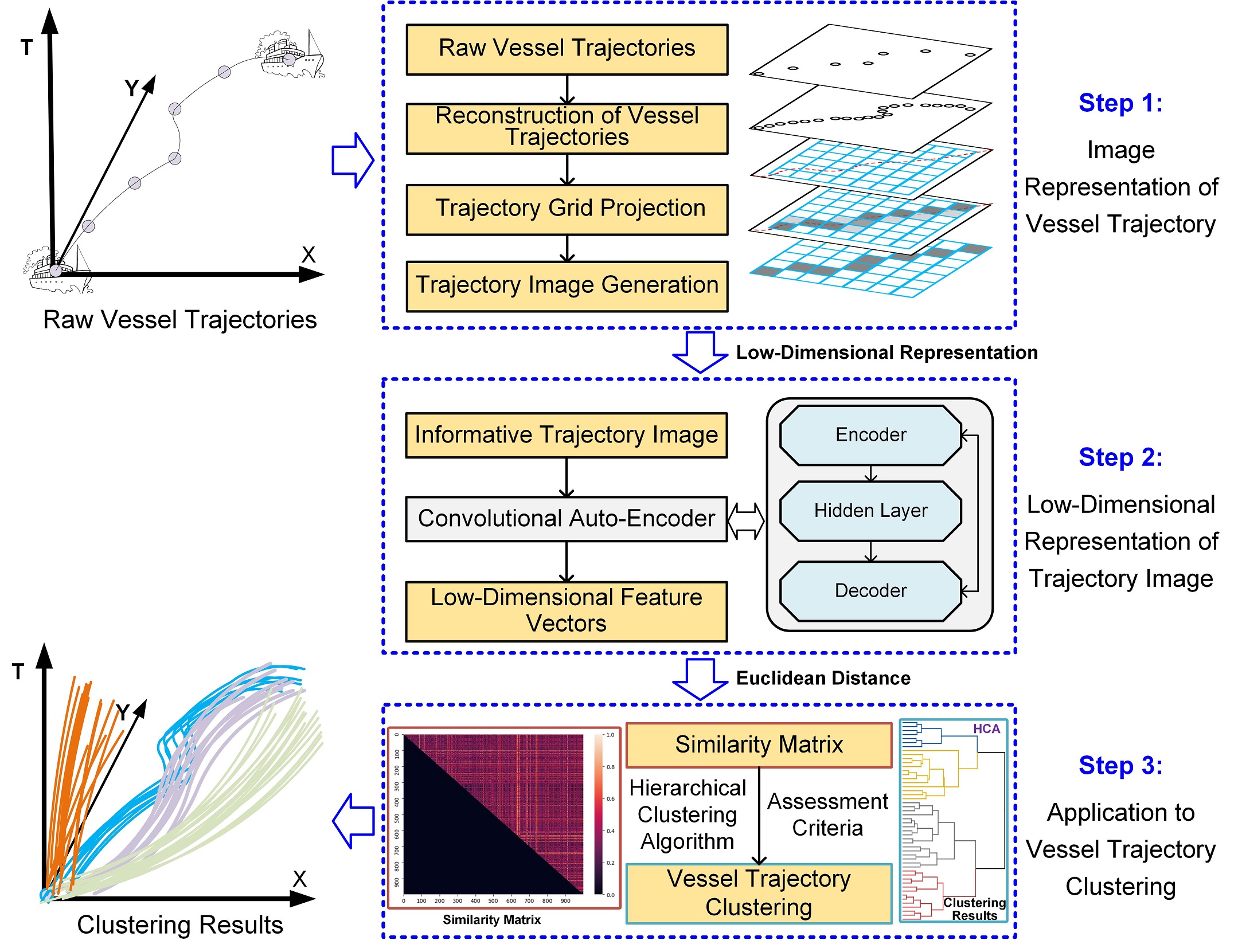}
	\caption{The flowchart of our proposed CAE-based unsupervised learning method for vessel trajectory similarity computation and its application to vessel trajectory clustering.}
	\label{fig:flowchart}
\end{figure}
\section{Introduction}
\label{sec:introduction}
The understanding of maritime traffic has become an important research topic due to human-related vessel activities and for its direct implication on maritime surveillance. Being mandatory by international regulations for vessels over $300$ gross tonnage and passenger vessels, the growing number of onboard automatic identification system (AIS) is providing massive spatio-temporal of vessel position data \citep{alessandrini2018estimated}. The collected AIS-based vessel trajectories are beneficial for vessel behavior modeling, leading to a reliable understanding of maritime traffic. In both academia and industry, vessel trajectory-based computational methods have gained significant attentions in maritime applications, such as abnormal behavior detection \citep{rong2020data}, maritime safety and security \citep{ou2008ais}, route planning \citep{zhang2018data}, fisheries and environment monitoring \citep{kroodsma2018tracking}, etc. These practical applications are strongly related to basic computing methods in vessel trajectory data mining. One of the most basic methods is vessel trajectory similarity computation, which is the main focus of this work.

In the current literature \citep{su2020survey}, there are several similarity computation methods with the goal to measure how similar any two trajectories are, such as the longest common subsequence (LCSS) \citep{yao2017trajectory}, edit distance \citep{fu2017finding}, one way distance (OWD) \citep{zhao2019novel}, and dynamic time warping (DTW) \citep{li2017dimensionality, de2012machine}, etc. These methods and their extensions have contributed to computing the similarities between vessel trajectories, assisting in learning vessel motion patterns from massive AIS-based vessel trajectories. However, these traditional methods often suffer from several drawbacks, such as high computational cost and sensitivity to unwanted noise and non-uniform sampling rate. In the case of large-scale vessel trajectories in maritime applications, it is intractable to efficiently measure the trajectory similarities or cluster trajectories \citep{taghizadeh2019meaningful}. The original vessel trajectories can be regarded as a combination of different sub-trajectories, which commonly have different geometrical structures. Therefore, these mentioned methods easily fail to generate high-quality similarity computation results. It is well known that deep learning has attracted impressive results in many different areas of research due to its strong feature representation \citep{lecun2015deep}. Inspired from the recent success of deep learning, we propose to develop an unsupervised learning method with convolutional auto-encoder (CAE) for vessel trajectory similarity computation. The proposed method has superior performance in terms of both efficiency and effectiveness.

Fig. \ref{fig:flowchart} visually depicts the flowchart of our CAE-based unsupervised learning method for computing similarities between vessel trajectories. To take full advantage of CAE in handling two-dimensional image data, we first transform the original vessel trajectories into informative trajectory images. The generated trajectory images are commonly robust to unwanted noise and non-uniform sampling rates existed in original trajectories. To accelerate similarity computation, we then propose to map the informative trajectory images onto the low-dimensional feature embedding through CAE model. The similarities between vessel trajectories are thus equivalent to efficiently calculating the distances between the low-dimensional feature vectors learned from the corresponding trajectory images. Last, the superior trajectory clustering results can also be guaranteed based on the accurate similarities calculated.

In conclusion, given the state-of-the-art research works, our CAE-based vessel trajectory similarity computation method obviously differs from previous scenarios in the following aspects
\begin{itemize}
	\item We propose to generate the informative trajectory images by remapping the original vessel trajectories into two-dimensional matrices. The similarities between different vessel trajectories are then equivalent to measuring the structural similarities between the corresponding informative trajectory images.
	\item A convolutional auto-encoder neural network is presented to learn low-dimensional representations of the informative trajectory images in an unsupervised manner. The learned representations could make vessel trajectory similarity more robust to trajectories with non-uniform sampling and different lengths.
	\item Comprehensive experiments on realistic data sets have demonstrated that the proposed method largely outperforms traditional vessel trajectory similarity computation methods in terms of efficiency and effectiveness.
\end{itemize}

The remainder of this paper is organized as follows. In Section \ref{sec:relatedworks}, we discuss the related works on trajectory similarity computation methods. Section \ref{sec:definitions} briefly reviews the definitions and preliminaries related to vessel trajectory, convolutional network and CAE. We then propose the CAE-based unsupervised learning method for vessel trajectory similarity computation in Section \ref{sec:trajsimilarity}. Section \ref{sec:results} performs extensive experiments to evaluate the superior performance of our proposed method. The conclusions and future work are finally summarized in Section \ref{sec:conclusion}.
%
%
\section{Related Works}
\label{sec:relatedworks}
Considerable research has been conducted on vessel trajectory similarity computation, which is of importance in maritime big data mining, such as navigation pattern discovery, trajectory classification, anomaly detection, and route planning \citep{tu2017exploiting}. In the literature \citep{besse2016review,su2020survey}, conventional similarity computation methods can be roughly divided into two categories, i.e., shape- and warping-based scenarios. The shape-based similarity measure methods commonly consider only geometrical shapes of trajectories. In contrast, the warping-based methods compute an optimal match between every two trajectories. Besides these two traditional methods, learning-based trajectory similarity computation methods have also attracted significant attention due to the powerful representation of auto-learning capacity. To our knowledge, no previous studies have been performed on deep learning-based similarity computation among vessel trajectories. There is thus a significant potential to promote the robustness, effectiveness and efficiency of trajectory similarity measure through deep learning-driven computation methods.
\subsection{Shape-Based Trajectory Similarity Computation}
The shape-based trajectory similarity computation methods typically include Hausdorff distance, Fr{\'e}chet distance, and OWD, etc. These methods have been widely adopted in maritime applications. For instance, \citet{arguedas2017maritime} proposed to compute the Hausdorff distance to cluster together similar vessel behaviours assisting in producing maritime traffic representations from historical AIS data. This distance has also been introduced to enable event identification in the maritime knowledge discovery \citep{sanchez2019simplification}. The OWD was adopted to assist in automatically recognizing the vessel motion patterns from massive historical AIS data \citep{ma2014vessel}. To enhance trajectory clustering performance, the Fr{\'e}chet distance has been used to enable adaptive vessel trajectory clustering \citep{cao2018pca}. It also contributed to developing a shape-based local spatial association measure for detecting abnormal activities of moving vessels \citep{roberts2019shape}. However, these shape-based methods are commonly sensitive to unwanted noise and non-uniform sampling rates. The related trajectory data mining results would be accordingly degraded in maritime applications.
\subsection{Warping-Based Trajectory Similarity Computation}
In current literature, several warping-based similarity computation methods, which perform well for shape matching in similarity measure, have been designed for trajectory data mining, such as DTW \citep{li2017dimensionality, de2012machine}, LCSS \citep{zhang2006comparison}, ED \citep{gustafsson2010particle}, edit distance on real sequences (EDR) \citep{de2012machine}, and edit distance with real penalty (ERP) \citep{chen2005robust}, etc. DTW has undoubtedly become one of the most popular methods to recognize similar geometrical patterns between different trajectories. It has achieved massive successes in the fields of vessel trajectory prediction \citep{alizadehvessel}, path-following autonomous vessels \citep{xu2019use}, shipping route characterization and anomaly detection \citep{rong2020data}. However, DTW-based similarity computation essentially suffers from over-stretching and over-compression problems. To eliminate these disadvantages, \citet{li2020adaptively} proposed an adaptively constrained DTW to robustly calculate the similarities and achieved reliable classification and clustering results. Other extensions of DTW, e.g., shape DTW \citep{zhao2018shapedtw}, shape segment DTW \citep{hong2020ssdtw}, and weighted DTW \citep{jeong2011weighted}, etc., could be naturally employed to measure similarities among vessel trajectories. Due to the recursive implementation and pair-wise distance computation, these methods were implemented at the expense of expensive computational costs. Both EDR and ERP have contributed to assisting in performing AIS-based vessel destination prediction \citep{zhang2020ais}. They still suffer from heavy computational burden in general, especially for large-scale vessel trajectories. To make trajectory data mining tractable in maritime applications, it is thus necessary to develop computational methods to efficiently measure trajectory similarities.
\subsection{Learning-Based Trajectory Similarity Computation}
With the progressive development of deep learning, several learning models, such as recurrent neural network (RNN) and convolutional neural network (CNN), have gained considerable success in trajectory similarity computation \citep{wang2019deep}. The reason behind this success attributes to their powerful feature learning abilities \citep{lecun2015deep}. Inspired by the recent success of RNN, \citet{yao2017trajectory} developed the RNN-based Seq2Seq auto-encoder model to encode the movement pattern of trajectory to improve similarity computation. By introducing a spatial proximity aware loss function, a Seq2Seq-based learning method (named t2vec) has been proposed to learn fixed-dimensional representations for trajectories \citep{li2018deep}. It could achieve significant similarity computation results which are robust to non-inform, low-sampling rates and random noisy data. The meaningfulness of learned similarity values in t2vec have been extensively assessed in the literature \citep{taghizadeh2019meaningful}. To overcome the influence of unwanted noise on similarity measure, \citet{zhang2019clustering} presented a robust auto-encoder model with attention mechanism to learn the low-dimensional representations of noisy vessel trajectories. \citet{zhang2019deep} further extended the Seq2Seq auto-encoder model (named At2vec) by considering the semantic information of active trajectories. The advantage of CNN architecture has been exploited to drive high-level features from raw GPS trajectories to infer transportation modes, potentially leading to traffic safety enhancement \citep{dabiri2018inferring}. Inspired by the huge successes of CNN and auto-encoder techniques, we will propose an unsupervised learning method with CAE to compute similarities between different vessel trajectories.
\section{Definitions and Preliminaries}
\label{sec:definitions}
In this section, we first introduce the preliminary concepts about vessel trajectories and challenging problems arising from trajectory similarity computation. Both AE and neural networks are then briefly reviewed.
\subsection{Definitions and Problem Statements}
The definitions of vessel trajectory and trajectory similarity are first given towards a better understanding of trajectory similarity computation. The core problems of this work are also introduced in this subsection, i.e., how to compute and evaluate the vessel trajectory similarity.
\begin{defn}
	\textbf{Vessel Trajectory}: A vessel trajectory $T$ is represented by a sequence of timestamped points collected from AIS equipment, i.e., $T = \left\{ P_{1}, P_{2}, \cdots, P_{N} \right\}$ where $P_{n} = \left\{ \mathrm{lat}_{n}, \mathrm{lng}_{n} , t_{n} \right\}$ with $n \in \{1, 2, \cdots, N  \}$ denoting the $n$-th timestamped point and $N$ being the length of vessel trajectory $T$. $\mathrm{lon}_{n}$, $\mathrm{lat}_{n}$ and $t_{n}$ in $P_{n}$, respectively, represent the longitude, latitude, and time stamp.
\end{defn}\label{def:VesselTraj}
\begin{defn}
	\textbf{Trajectory Similarity}: The trajectory similarity is commonly measured by computing the distance $\mathcal{D} (T_{i}, T_{j})$ between any two trajectories $T_{i}$ and $T_{j}$ with $i \neq j$. Theoretically, the smaller the value $\mathcal{D}$ is, the higher similar the two trajectories are.
\end{defn}\label{def:TrajSimilar}

In maritime applications, it is more intuitive and tractable to compute the distances (rather than similarities) between different vessel trajectories. The trajectory similarity measure can thus be naturally equivalent to computing the distance. For any three trajectories $T_1$, $T_2$ and $T_3$, the distance measure $\mathcal{D}$ should meet the following four conditions
\begin{enumerate}
	\item Symmetry: $\mathcal{D} (T_{1}, T_{2}) = \mathcal{D} (T_{2}, T_{1})$.
	\item Non-negativity: $\mathcal{D} (T_{1}, T_{2}) \geq 0$.
	\item Triangle inequality: $\mathcal{D} (T_{1}, T_{3}) \leq \mathcal{D} (T_{1}, T_{2}) + \mathcal{D} (T_{2}, T_{3})$.
	\item Identity of indiscernibles: $\mathcal{D} (T_{1}, T_{2}) = 0 \Leftrightarrow T_1 = T_2$.
\end{enumerate}

A large number of computation methods have been proposed to measure trajectory similarity \citep{besse2016review,su2020survey}, it is still, however, difficult to generate satisfactory measure performance in terms of efficiency, effectiveness and robustness. As a result, traditional methods thus often fail to yield robust scenarios in realistic trajectory datasets. To alleviate these disadvantages, it is necessary to handle the following two challenging problems
\begin{prob}
	\textbf{How to Efficiently Compute the Trajectory Similarity?}: The trajectory similarity computation is strongly dependent upon the geometrical properties of trajectories. Traditional shape- and warping-based similarity computation methods easily suffer in several scenarios, e.g., long computational time, unequal lengths, unwanted artifacts (noise), non-uniform sampling rates, etc. To guarantee efficiency and accuracy, we will propose a unsupervised feature learning method to enable computation of vessel trajectory similarity. The proposed learning method is capable of significantly reducing the computational cost while accurately measuring the trajectory similarity.
\end{prob}
\begin{prob}
	\textbf{How to Robustly Evaluate the Trajectory Similarity?}: If any two trajectories have the same length and time stamps, it becomes significantly easier to evaluate the trajectory similarity. However, unwanted artifacts, unequal lengths and different time stamps are more common, leading to intractable evaluation of trajectory similarity in practice. To our knowledge, there is no golden standard or benchmarks to quantitatively evaluate the similarity measure thus far. To overcome this limitation, we propose to adopt the trajectory clustering results to indirectly evaluate the trajectory similarity computation.
\end{prob}
\begin{figure}[t]	
	\centering\includegraphics[width=0.8\linewidth]{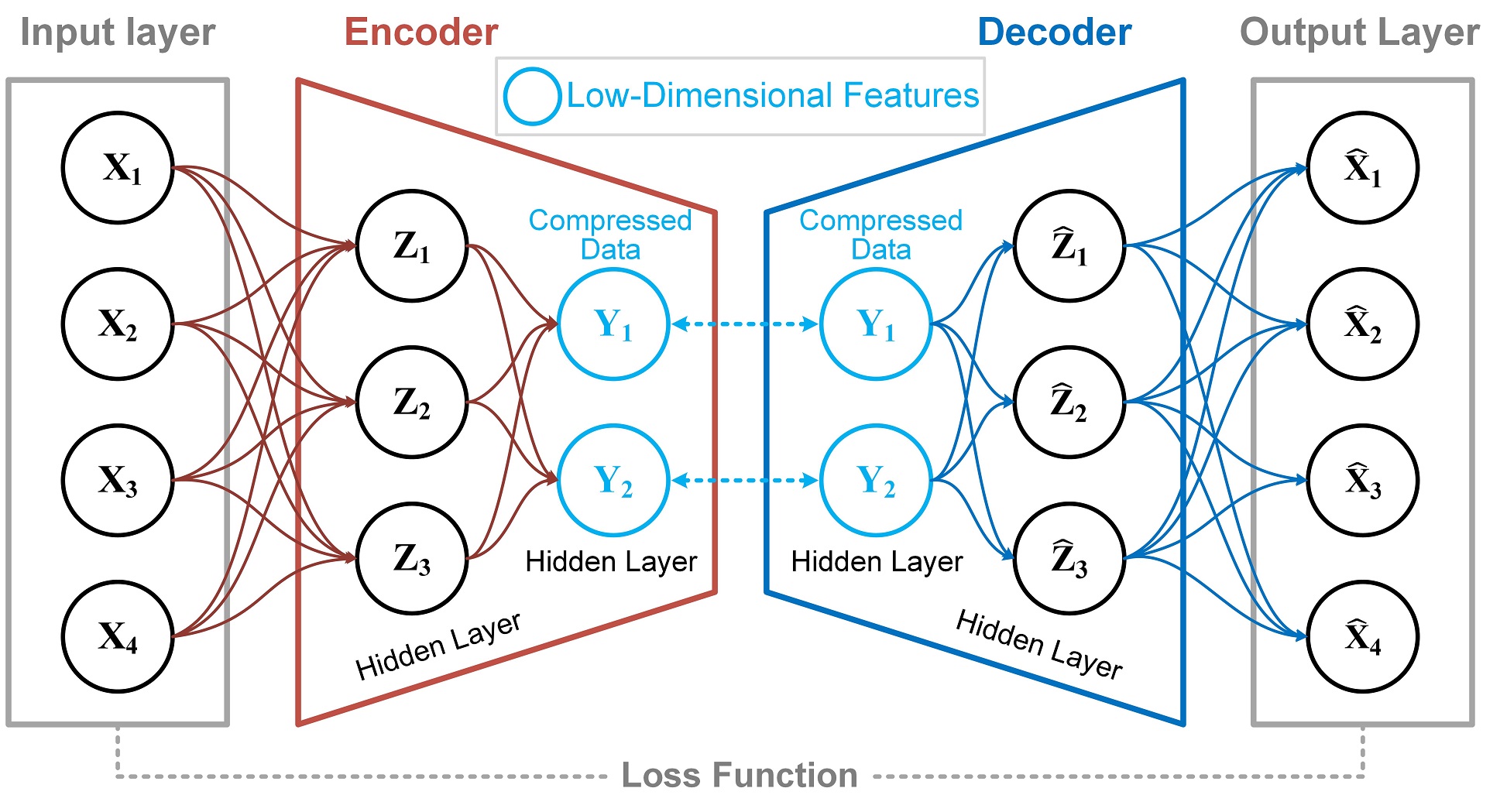}
	\caption{The architecture of a traditional auto-encoder model. It is an unsupervised learning method where the purpose of encoder layer is to learn a low-dimensional representation (i.e., feature or compressed data) of the training example. In contrast, the decoder maps the low-dimensional feature to the reconstruction of original input.}
	\label{fig:AE}
\end{figure}
\subsection{Auto-Encoder Neural Network}

The basic auto-encoder (AE) is a kind of unsupervised neural network, which aims to encode input samples into the low-dimensional representations. The inputs in AE can be effectively reconstructed from these representations with minimum reconstruction error. The architecture of a traditional AE is visually illustrated in Fig. \ref{fig:AE}. It can be observed that AE is mainly composed of an input layer, several hidden layers, and an output layer. Let $x_{m} = \left[ x_{m}^{1}, x_{m}^{2}, \cdots, x_{m}^{P} \right]^{T}$ with $1 \leq m \leq M$ be an unlabeled training sample, where $P$ and $M$ denote the length of one sample $x_{m}$, and the number of unlabeled samples, respectively. The first step of AE is to encode the input data $x_{m}$ to the compressed data (i.e., low-dimensional features) $y_{m} = \left[ y_{m}^{1}, y_{m}^{2}, \cdots, y_{m}^{Q} \right]^{T}$ with $Q \ll P$ through the following activation function
\begin{equation}\label{eq:encoder}
    y_{m} = \mathcal{A}_{\mathcal{E}} \left( \mathcal{W} x_{m} + \mathcal{B} \right),
\end{equation}
where $\mathcal{W}$ is the weight matrix, $\mathcal{B}$ is a bias vector, and $\mathcal{A}_{\mathcal{E}} \left( \cdot \right)$ denotes the activation function. The next step of AE is to reconstruct the $P$-dimensional vector $\hat{x}_{m} = \left[ \hat{x}_{m}^{1}, \hat{x}_{m}^{2}, \cdots, \hat{x}_{m}^{P} \right]^{T}$ based on the low-dimensional features $y_{m}$, i.e.,
\begin{equation}\label{eq:decoder}
    \hat{x}_{m} = \mathcal{A}_{\mathcal{D}} \left( \hat{\mathcal{W}} y_{m} + \hat{\mathcal{B}} \right),
\end{equation}
with the appropriately sized parameters $\hat{\mathcal{W}}$ and $\hat{\mathcal{B}}$. Here, $\mathcal{A}_{\mathcal{D}} \left( \cdot \right)$ also denotes the activation function. The essential purpose of AE training is to optimize the parameter set $\Theta = \{ \mathcal{W}, \mathcal{B}, \hat{\mathcal{W}}, \hat{\mathcal{B}} \}$ to minimize the reconstruction error. Generally, the reconstruction error between input and output data is commonly defined as the mean-squared error (MSE) cost function, i.e.,
\begin{equation}\label{eq:costfunction}
    \mathcal{E} \left( \Theta \right) = \frac{1}{M} \sum_{m=1}^{M} \left( \frac{1}{2} \sum_{p=1}^{P} \left( {x}_{m}^{p} - \hat{x}_{m}^{p} \right)^{2} \right).
\end{equation}

It is computationally tractable to minimize the reconstruction error (\ref{eq:costfunction}) by optimizing the parameter set $\Theta = \{ \mathcal{W}, \mathcal{B}, \hat{\mathcal{W}}, \hat{\mathcal{B}} \}$ in both encoder and decoder layers to yield the low-dimensional features. These obtained features are beneficial for dimensionality reduction leading to reduced computational complexity in many application scenarios.
\begin{figure}[t]	
	\centering\includegraphics[width=1\linewidth]{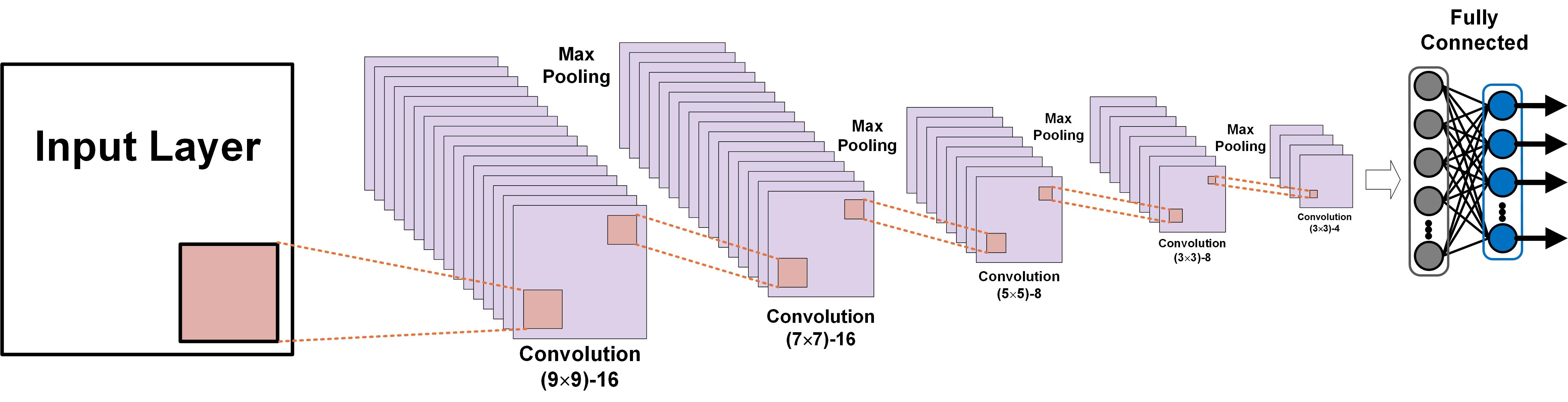}
	\caption{The visual architecture of a typical convolutional neural network (CNN).}
	\label{fig:CNN}
\end{figure}
\subsection{Convolutional Neural Network}
CNN can be regarded as a feed-forward artificial neural network with variations of multilayer perceptrons \citep{diaz2018modelling}. It has been extensively exploited in several practical fields, such as machine vision, computer graphics, speech recognition, and big data mining, etc \citep{zhang2018survey,khan2020survey}. These successful applications are obtained mainly due to the immense learning capacity of CNN. This capacity can enable automatic extraction of multiple features leading to meaningful representations in different abstraction levels. The overall architecture of a typical CNN is visually shown in Fig. \ref{fig:CNN}. In particular, the conventional CNN structure is mainly composed of several convolutional layers, pooling layers, fully-connected layers, and some activation functions, which will be briefly reviewed next.
%
%
\subsubsection{Convolutional Layers}
The convolutional layers essentially work as feature extractors to learn the rich feature representations of input samples, which are two-dimensional images in this work. The convolutional filters within convolutional layers implement via dividing the input images into several small patches, often known as receptive fields. These filters mainly have two important parameters (i.e., height and weight), which are commonly smaller than those of the input images. The receptive fields related to the input images are convolved with filters using a set of trainable weights to generate new feature maps. In particular, each feature map is yielded by sliding the filter over all spatial locations of the input. The dimensions of complete feature maps are equivalent to the number of different filters used. Mathematically, the feature value $x_{k}^{l}$ in the $k$-th feature map of the $l$-th convolutional layer is computed as follows
\begin{equation}
    x_{k}^{l} = f_{k}^{l} \odot x_{k}^{l-1} + b_{k}^{l},
\end{equation}
where $\odot $ represents the two-dimensional convolutional operator, $f_{k}^{l}$ is the $k$-th convolutional filter in the $l$-th layer, $x_{k}^{l-1}$ denotes the input patch for the $k$-th filter within the $(l - 1)$-th layer, and $b_{k}^{l}$ is the bias term related to the $k$-th filter of the $l$-th layer. The convolutional filter in traditional CNN is a generalized linear expression. To enhance the representation ability of features, there have been several generalizations and extensions, e.g., tiled convolution, transposed convolution, and dilated convolution, etc \citep{gu2018recent}.
\subsubsection{Activation Functions}
Activation functions are important concepts in building deep neural networks, which determine how networks learn and behave. The activation can be selected as an arbitrary function which transforms the output of a layer into one network. In particular, the activation function introduces non-linearity to networks and makes them capable of learning and performing more complex tasks. Without activation functions, the neural networks are essentially just linear regression estimators. Let $x_{k}^{l}$ be an output (i.e., convolved feature) of the $k$-th filter in the $l$-th convolutional layer, a non-linear activation function $\mathcal{A} \left( \cdot \right)$ for the convolved feature can be defined as follows
\begin{equation}\label{eq:activationfunction}
    v_{k}^{l} = \mathcal{A} \left( x_{k}^{l} \right),
\end{equation}
where $v_{k}^{l}$ denotes the transformed output of the $k$-th feature map for the $l$-th convolutional layer. The above-mentioned functions $\mathcal{A}_{\mathcal{E}} \left( \cdot \right)$ and $\mathcal{A}_{\mathcal{D}} \left( \cdot \right)$ in Eqs. (\ref{eq:encoder}) and (\ref{eq:decoder}) can be selected as the same versions of $\mathcal{A} \left( \cdot \right)$ in this work.

In current literature, there have been attempts to engineer several activation functions $\mathcal{A} \left( \cdot \right)$ to improve neural networks, such as Sigmoid \citep{iliev2017approximation}, Hyperbolic Tangent \citep{dabiri2018inferring}, Rectified Linear Unit (ReLU) \citep{yao2017trajectory}, Parametric ReLU (PReLU) \citep{he2015delving}, Randomized ReLU (RReLU) \citep{yao2017trajectory}, Leaky ReLU \citep{anthimopoulos2016lung}, and Exponential Linear Unit (ELU) \citep{clevert2015fast}, etc. Note that different activation functions often display diverse behaviors in practical applications. It is thus necessary to select the proper activation function to enable fast and robust training of neural networks.
\subsubsection{Pooling Layers}
The pooling layers are usually placed between two successive convolutional layers. Their purposes are to generate translation and distortion invariance in the input data by down-sampling the feature maps. It is able able to summarize similar information within the neighborhoods of receptive fields and yield the dominant response within these local regions. Let $\mathrm{pool} \left( \cdot \right)$ denote the pooling function, the pooled element in each feature map is then given by
\begin{equation}\label{eq:poolinglayer}
    w_{k}^{l} = \mathrm{pool} \left( v_{k}^{l} \right),
\end{equation}
where $w_{k}^{l}$ denotes the pooled feature map $v_{k}^{l}$. The typical pooling operations mainly include max pooling and average pooling \citep{gu2018recent}. The reduced spatial resolution of feature maps not only regulates the computational complexity but also improves the generalization of learning by reducing overfitting. In current literature \citep{gu2018recent,dakhia2019multi}, several extensions of pooling operators, such as $L_{p}$ pooling, mixed pooling, stochastic pooling, spectral pooling, spatial pyramid pooling, and multi-scale orderless pooling, have been proposed to further enhance the generalization ability of CNN.
\subsubsection{Fully-Connected Layers}
Typically, one CNN commonly has several convolutional and pooling layers, which extract meaningful feature representations, followed by one or more fully-connected layers. The fully-connected layers are able to interpret the feature representations and perform the high-level reasoning. In addition, these layers essentially perform a non-linear combination of features learned from previous layers, which are beneficial for classification of experimental samples.
\section{CAE-Enabled Trajectory Similarity Computation Framework}
\label{sec:trajsimilarity}
As discussed beforehand, CNN is able to quickly extract meaningful features from input examples. In contrast, the unsupervised AE is capable of encoding input examples into the low-dimensional representations. To take full advantages of CNN and AE, there is a great potential to combine CNN and AE contributing to CAE neural network. In this work, the raw vessel trajectories are firstly remapped into informative trajectory images. CAE will then be adopted to learn the low-dimensional features related to the informative trajectory images. The similarities between different vessel trajectories can be equivalent to computing the distances between the corresponding low-dimensional feature vectors. The proposed CAE-based unsupervised learning method could achieve superior similarity measures in terms of both efficiency and effectiveness.
\begin{figure}[t]
	\centering\includegraphics[width=1\linewidth]{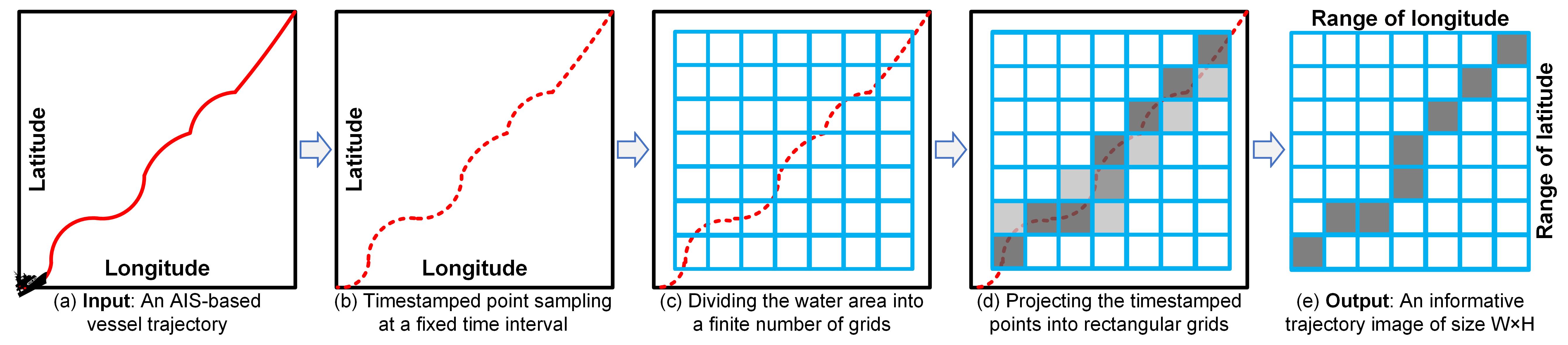}
	\caption{The visual overview of generation of informative trajectory image. From left to right: (a) one raw vessel trajectory, (b) timestamped point sampling using the cubic spline interpolation at a fixed time interval, (c) dividing the experimental water area into a finite number of grids, (d) projecting the timestamped points in the raw trajectory into rectangular grids, and (e) the corresponding informative trajectory image of size $W \times H$, respectively.}
	\label{fig:TrajImage}
\end{figure}
\subsection{Representation of Raw Vessel Trajectories}
To take full advantage of CAE, it is necessary to firstly transform raw vessel trajectories into two-dimensional images. As shown in Fig. \ref{fig:TrajImage}, we propose to generate the informative trajectory images $\mathcal{X}$ by projecting the timestamped points in trajectories $T$ into the predefined grids. In this work, we propose to select the Cartesian coordinates to represent the geometric positions of timestamped points through the Mercator projection \citep{li2016trajectory}. The generation of informative trajectory images is mainly composed of three steps. In the first step, we propose to resample the timestamped points in the original vessel trajectories at a fixed time interval of $5$s. If the raw trajectories are degraded by random noise or missing data, the spline interpolation method \citep{zhang2018novel} will be selected to reconstruct the degraded timestamped points leading to trajectory quality improvement. The water areas considered in this work will then be divided into a finite number of grids of size $W \times H$ in the second step. In the last step, the timestamped points in vessel trajectories are projected into the grids leading to the informative trajectory images of size $W \times H$. In this work, the trajectory images are simplistically considered as binary matrices containing only $0$s and $1$s. Before generating the binary informative trajectory images $\mathcal{X}$, we will first compute the projected images $\mathcal{J}$ by analyzing the timestamped points projected.

Let $T = \left\{ P_{1}, P_{2}, \cdots, P_{N} \right\}$ denote a vessel trajectory with $P_{n} = \left\{ \mathrm{lat}_{n}, \mathrm{lng}_{n} , t_{n} \right\}$, where $n \in \{1, 2, \cdots, N  \}$ denotes the $n$-th timestamped point and $N$ is the length of $T$ defined in Definition \ref{def:VesselTraj}. The mathematical relationship between $\left( w_{n}, h_{n} \right)$ in projected image $\mathcal{J}$ and $\left( \mathrm{lat}_{n}, \mathrm{lng}_{n} \right)$ in Cartesian coordinates \citep{huang2020gpu} can be obtained as follows
\begin{equation}
    w_{n} = \lceil \frac{ \mathrm{lat}_{n} - \mathrm{lat}_{\min} }{\mathrm{lat}_{\max} - \mathrm{lat}_{\min}} \cdot \left( W - 1 \right) \rceil + 1 \in \left[ 1, W \right],
\end{equation}
with $\mathrm{lat}_{\min} = \min_{1 \leq n \leq N} \mathrm{lat}_{n}$, $\mathrm{lat}_{\max} = \max_{1 \leq n \leq N} \mathrm{lat}_{n}$, and
\begin{equation}
    h_{n} = \lceil \frac{ \mathrm{lng}_{n} - \mathrm{lng}_{\min} }{\mathrm{lng}_{\max} - \mathrm{lng}_{\min}} \cdot \left( H - 1 \right) \rceil + 1 \in \left[ 1, H \right],
\end{equation}
with $\mathrm{lng}_{\min} = \min_{1 \leq n \leq N} \mathrm{lng}_{n}$, $\mathrm{lng}_{\max} = \max_{1 \leq n \leq N} \mathrm{lng}_{n}$, and $\lceil \cdots \rceil$ denoting the ceil operation. If the number of timestamped points, whose projected coordinates are $\left( w_n, h_n \right)$, equals $d_n$, the intensity of element $\left( w_n, h_n \right)$ in $\mathcal{J}$ becomes $\mathcal{J} \left( w_n, h_n \right) = d_n$ essentially indicating the magnitude of projected image $\mathcal{J}$. In this work, we can adopt the intermediate image $\mathcal{J}$ to generate the binary trajectory image $\mathcal{X}$, i.e.,
\begin{equation}
\mathcal{X} \left( w_n, h_n \right) =
  \begin{cases}
    1, & \text{if}~\mathcal{J} \left( w_n, h_n \right) > \epsilon, \\
    0, & \text{otherwise},
  \end{cases}
\end{equation}
where $\epsilon$ is a predefined threshold, empirically selected as $\epsilon = 3$ in this work. Experimental results have demonstrated the effectiveness of this brief selection.
\begin{figure}[t]
	\centering\includegraphics[width=1\linewidth]{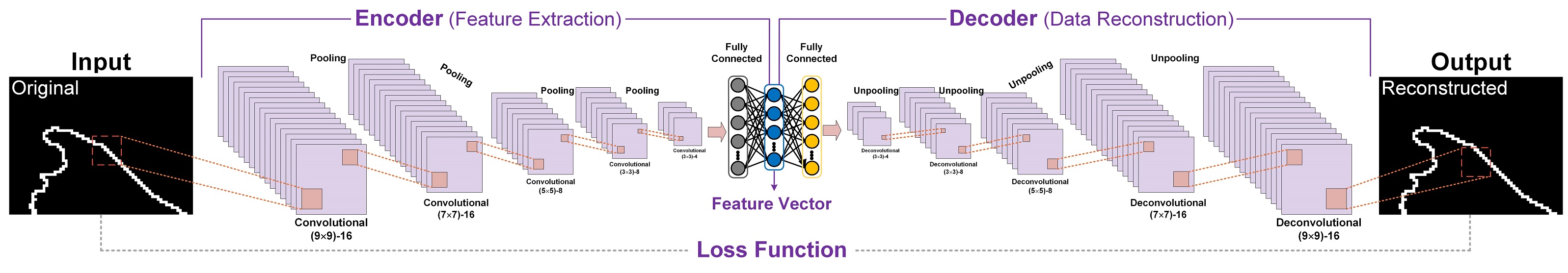}
	\caption{The architecture of CAE neural network. The CAE mainly consists of encoder and decoder components, connected at the narrowest location of the network. The loss function between input data and its reconstructed version is adopted to optimize the weights and bias in CAE neural network.}\label{fig:CAEstructure}
\end{figure}
\subsection{CAE-Based Unsupervised Learning Method}
\label{sec:CAEmodel}
To fully extract features from the informative trajectory images, we propose a multi-layer CAE neural network to perform unsupervised similarity learning. The proposed unsupervised learning method essentially combines the advantages and architectures of CNN and AE. In particular, the CNN is capable of learning local features of input data due to its spatially located convolutional filters, whereas the AE is able to directly extract global features while not requiring labeled data. To extract the localized features, the convolutional layers, where learnable weights are shared among all locations in previous layers, are incorporated into AE contributing to the implementation of CAE. As shown in Fig. \ref{fig:CAEstructure}, the encoder component of CAE is responsible for transforming the original informative trajectory images into low-dimensional feature vectors. The decoder layers are trained to effectively and robustly reconstruct the informative trajectory images from the learned features. The detailed architecture of our CAE-based unsupervised learning method will be described next.
\begin{figure}[t]
	\centering\includegraphics[width=1\linewidth]{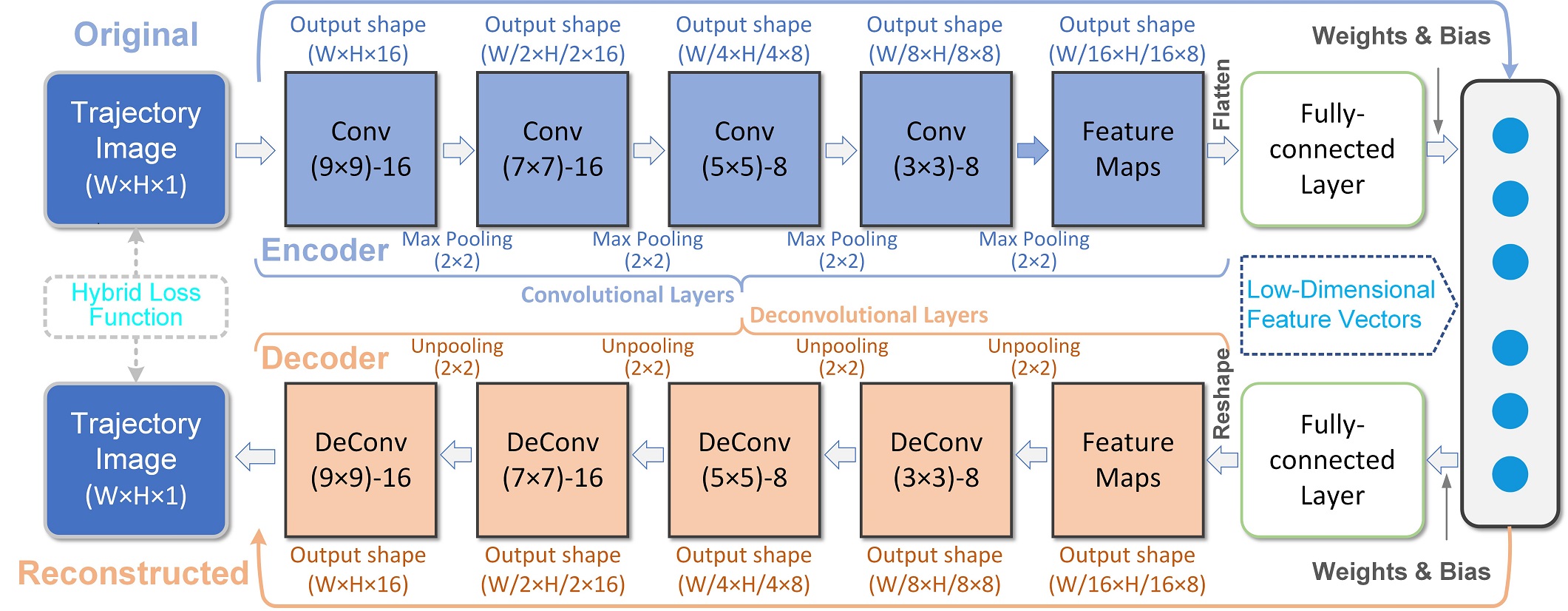}
	\caption{The visual architecture of our CAE-based unsupervised learning method, which mainly consists of the convolutional-deconvolutional auto-encoder and fullly-connected layers. The parameters of each layer are denoted by "(pooling size)" for max pooling and unpooling layers, and "(filter size)-(number of filters)" for Conv and DeConv layers. The output sizes of both convolutional and deconvolutional layers are represented by "Output shape".}
	\label{fig:CAEour}
\end{figure}
\subsubsection{Network Architecture}
The architecture of our CAE neural network is visually illustrated in Fig. \ref{fig:CAEour}. The proposed CAE-based learning method mainly consists of fully-connected layers, encoder and decoder components. The input of CAE is the informative trajectory image transformed from one specific vessel trajectory. The encoder component mainly consists of $4$ convolutional layers and $4$ max-pooling layers followed by one fully-connected layer as the last layer. The feature map in each convolutional layer is generated by performing convolution between the input (or feature map from the previous layer) and the learnable filter, and max-pooling operation. In particular, the sizes and numbers of filters in these convolutional layers are set to $9 \times 9$, $7 \times 7$, $5 \times 5$, $3 \times 3$, and $16$, $16$, $8$, $8$, respectively. The $2 \times 2$ max-pooling layer is introduced to calculate the maximum value of every $2 \times 2$ region in each feature map. Therefore, the size of feature map in current layer will reduce by a half size of previous layer only after max-pooling operation. The convolved features are activated through the widely-used ReLU function $\mathcal{A}_{\mathcal{E}} \left( \cdot \right)$. The main purpose of encoder component is to generate meaningful local feature maps of size $\frac{W}{16} \times \frac{H}{16} \times 8$. To learn the low-dimensional representations of raw vessel trajectories, the fully-connected neural networks, which match the size of final feature map, are adopted to generate the one-dimensional feature vector $Y$ of length $L$ in this work\footnote{We will detailedly discuss how to optimally select the length $L$ in Section \ref{sec:DimFeature}.}. For the sake of simplicity, we generally calculate the feature map in the next layer through the following equation
\begin{equation}
    x_{k}^{l} = \mathcal{A}_{\mathcal{E}} \left( f_{k}^{l} \odot x_{k}^{l-1} + b_{k}^{l} \right),
\end{equation}
with $1 \leq l \leq 4$ and $x^{0} = \mathcal{X}$. Here, $\odot$ denotes the convolutional operator, $f_{k}^{l}$ is the convolutional filter, $x_{k}^{l-1}$ denotes the feature map for the $k$-th filter within the $(l - 1)$-th layer, and $b_{k}^{l}$ is the bias term. The vectorial feature map $X$ of size $\frac{WH}{32} \times 1$ can be obtained by flattening the feature map tensor of size $\frac{W}{16} \times \frac{H}{16} \times 8$, i.e., $X = \mathrm{Flatten} \left( x_{k}^{4} \right)$. Therefore, the final one-dimensional feature vector $Y$, extracted from the informative trajectory image $\mathcal{X}$, can be obtained as follows
\begin{equation}
    Y = \mathcal{H} \left( \mathcal{X} \right) = \mathcal{W} X + \mathcal{B},
\end{equation}
where $\mathcal{H} \left( \cdot \right)$ can be regarded as the feature extraction operator, $\mathcal{W}$ and $\mathcal{B}$, respectively, denote the weight matrix and bias vector in fully-connected neural networks. Note that the compressed representation $Y$ commonly includes more useful properties than the original trajectory image $\mathcal{X}$. It is thus feasible to perform trajectory similarity computation only using this one-dimensional feature representation.

In the decoder component, we have the same number of layers and implement the inverse operations (i.e., unpooling and deconvolutional). It is thus able to yield an output (i.e., reconstructed version) with the same size of the input in the first layer of the encoder component. To reconstruct the input data $\mathcal{X}$, the one-dimensional feature vector $Y$ should be firstly transformed to the vectorial feature map $\hat{X}$ through fully-connected neural networks, i.e.,
\begin{equation}
    \hat{X} = \hat{\mathcal{W}} Y + \hat{\mathcal{B}},
\end{equation}
with $\hat{\mathcal{W}}$ and $\hat{\mathcal{B}}$ representing the weight matrix and bias vector, respectively. By reshaping the vectorial data $\hat{X}$ of size $\frac{WH}{32} \times 1$, we can obtain the feature map tensor $\hat{x}_{k}^{1}$ of size $\frac{W}{16} \times \frac{H}{16} \times 8$, i.e., $\hat{x}_{k}^{1} = \mathrm{Reshape} \left( \hat{X}, \left[ W/16, H/16, 8 \right] \right)$ with $\mathrm{Reshape} \left( \cdot \right)$ being a reshaping operation. The feature map in the $l$-th deconvolutional layer is defined as
\begin{equation}
    \hat{x}_{k}^{l} = \mathcal{A}_{\mathcal{D}} \left( \hat{f}_{k}^{l} \odot \hat{x}_{k}^{l-1} + \hat{b}_{k}^{l} \right),
\end{equation}
where $\mathcal{A}_{\mathcal{D}} \left( \cdot \right)$ is the ReLU activation function, $\hat{f}_{k}^{l}$ is the deconvolutional filter, and $\hat{b}_{k}^{l}$ is the bias term in deconvolutional layer. Finally, the reconstruction of input $\hat{\mathcal{X}}$ can thus be generated by linearly combining these estimated feature maps, i.e, $\hat{\mathcal{X}} = \sum_{k} \hat{x}_{k}^{l}$ with $l=4$ in the decoder component. Meanwhile, the $2 \times 2$ unpooling layer is adopted to restore the location information unsampled in pooling layers, shown in Fig. \ref{fig:CAEour}.
\subsubsection{Hybrid Loss Function}
The proposed CAE-based neural network is inevitably trained using an optimization process which requires a loss function to compute the model error. Mean squared error has become the most widely-used loss function in the fields of computer vision and data mining. However, this $L_{2}$-norm loss function is essentially sensitive to random outliers. In addition, the reconstructed results $\hat{\mathcal{X}}$ easily suffer from blurred details leading to unstable feature learning from informative trajectory images. To overcome this limitation, we propose to adopt $L_{1}$-norm loss function instead of $L_{2}$-norm version to minimize the sum of the absolute differences between original and reconstructed trajectory images. Let $\mathcal{X}_{m}$ and $\hat{\mathcal{X}}_{m}$, respectively, denote the $m$-th informative trajectory image and its reconstructed version. In this work, the definition of $L_{1}$-norm loss function is then given by
\begin{equation}\label{eq:L2loss}
	\mathcal{F}_{1} \left( \textbf{X}, \hat{\textbf{X}} \right) = \frac{1}{M} \sum_{m = 1}^{M} \left\| \mathcal{X}_{m} - \hat{\mathcal{X}}_{m} \right\|_{1},
\end{equation}
where $\left\| \cdot \right\|_{1}$ denotes the $L_{1}$-norm, $\textbf{X} = \{ \mathcal{X}_1, \mathcal{X}_2, \cdots, \mathcal{X}_M \}$, and $M$ is the total number of informative trajectory images projected from the raw vessel trajectories.

To further improve the feature learning ability of our CAE network, we introduce the perceptually-sensitive loss function, i.e., structural similarity (SSIM), to complement $\mathcal{F}_{1}$ in Eq. (\ref{eq:L2loss}). By taking into consideration the luminance, contrast and structure terms, SSIM is sensitive to structural information and texture, which shows good agreement with human observers. It enables the CAE model to learn more details from informative trajectory images, resulting in more robust trajectory similarity computation. The SSIM between $\mathcal{X}_{m}$ and $\hat{\mathcal{X}}_{m}$ is defined as follows
\begin{equation}
    \mathrm{SSIM} \left( \mathcal{X}_{m}, \hat{\mathcal{X}}_{m} \right) = \frac{2 \mu_{\mathcal{X}_{m}} \mu_{\hat{\mathcal{X}}_{m}} + c_{1} }{ \mu_{\mathcal{X}_{m}}^{2} + \mu_{\hat{\mathcal{X}}_{m}}^{2} + c_{1}} \cdot \frac{2 \sigma_{ \mathcal{X}_{m} \hat{\mathcal{X}}_{m} } + c_{2}} {\sigma_{\mathcal{X}_{m}}^{2} + \sigma_{\hat{\mathcal{X}}_{m}}^{2} + c_{2}},
\end{equation}
with $\mu_{\circ}$, $\sigma_{\circ}$, and $\sigma_{\circ, \circ}$ denoting the mean, standard deviation, and cross correlation, respectively. Both positive constants $c_1$ and $c_2$ are used to guarantee numerical stability of the division. Please refer to \citep{wang2004image} for more details on SSIM. The SSIM-based loss function is thus given by
\begin{equation}\label{eq:SSIM}
    \mathcal{F}_{2} \left( \textbf{X}, \hat{\textbf{X}} \right) = 1 - \frac{1}{M} \sum_{m = 1}^{M} \mathrm{SSIM} \left( \mathcal{X}_{m}, \hat{\mathcal{X}}_{m} \right),
\end{equation}

Therefore, the hybrid loss function $\mathcal{F}$ adopted in this work is formulated as follows
\begin{equation}\label{eq:lossfunction}
    \mathcal{F} = \lambda_{1} \mathcal{F}_1 + \lambda_{2} \mathcal{F}_2,
\end{equation}
where $\lambda_{1}$ and $\lambda_{2}$ are weighting parameters. These two parameters play important roles in CAE training related to trajectory similarity computation. We empirically set the weighting parameters to $\lambda_{1} = 0.15$ and $\lambda_{2} = 0.85$ which can guarantee robust learning performance in our experiments.

During the training procedure of our CAE, the back-propagation algorithm adjusts the weights and bias in the direction of the negative gradient of the hybrid loss function. Its purpose is to minimize the difference between the input data and its reconstructed version. Once the CAE neural network is trained well, only the encoder component will be deployed to provide the low-dimensional feature vectors in this work. These learned features are sufficient to assist in measuring the similarities between different vessel trajectories. It is worth noting that the other decoder component is only adopted to evaluate the reliability of CAE-based low-dimensional representation of informative trajectory images. Experimental results have shown that our unsupervised learning method could offer considerable advantages in terms of its practicality, robustness, and flexibility.
\subsection{Extension to Vessel Trajectory Clustering}
As discussed above, original vessel trajectories with massive timestamped points can be effectively represented by feature vectors with few elements. It means that the significant geometrical structures in original trajectories can be accurately reconstructed using the low-dimensional feature vectors. The similarities between different vessel trajectories are equivalent to efficiently computing the distances between the corresponding feature vectors, which are yielded through our CAE-based unsupervised learning method. Once the similarities or distances are obtained with low computational cost, it will be accordingly fast to perform clustering of vessel trajectories. The clustering performance is essentially depended upon the computation of trajectory similarity. The similarity computation results can thus be evaluated through the trajectory clustering results.

Let $\mathcal{M}$ denote a distance matrix storing the distance between any two trajectories. In particular, the distance $\mathcal{D}_{\mathcal{E}}$, inversely proportional to similarity, between the $i$- and $j$-th trajectories is defined as follows
\begin{equation}
    \mathcal{M}_{i,j} = \mathcal{D}_{\mathcal{E}} \left( \mathcal{H} \left( \mathcal{X}_{i} \right), \mathcal{H} \left( \mathcal{X}_{j} \right) \right) \leftarrow \mathcal{D} \left( T_{i}, T_{j} \right),
\end{equation}
with $i \neq j$. Here, the informative trajectory image $\mathcal{X}_{i}$ is projected from the $i$-th vessel trajectory $T_{i}$. It is easy to obtain $\mathcal{M}_{i,j} = 0$ if and only if $i = j$ in theory. In our experiments, we tend to adopt the simple Euclidean distance, i.e., $\mathcal{D}_{\mathcal{E}} \left( \mathcal{H} \left( \mathcal{X}_{i} \right), \mathcal{H} \left( \mathcal{X}_{j} \right) \right) = \sqrt{\sum_{l=1}^L ( h_{i}^{l} - h_{j}^{l} )^2}$ with $\mathcal{H} \left( \mathcal{X}_{i} \right) = \{ h_{i}^{1}, h_{i}^{2}, \cdots, h_{i}^{L}\}$, $\mathcal{H} \left( \mathcal{X}_{j} \right) = \{ h_{j}^{1}, h_{j}^{2}, \cdots, h_{j}^{L}\}$, and $L$ denoting the length of feature vector $\mathcal{H} \left( \circ \right)$.

Once the distance matrix $\mathcal{M}$ is obtained, several clustering algorithms \citep{fahad2014survey} can be adopted to classify each vessel trajectory into a specific group. Without loss of generality, we propose to directly select the widely-used hierarchical clustering algorithm (HCA) to perform vessel trajectory clustering in this work. Please refer to \citep{fan2019ope} for more details on HCA. If the similarity (i.e., distance) computing results are accurate and reliable, it becomes tractable to accurately cluster the vessel trajectories in maritime applications. Note that clustering operation could produce results that are more intuitive compared with similarity computing. Therefore, the trajectory clustering performance can contribute to qualitatively evaluating the effectiveness of trajectory similarity computation.
\section{Experimental Results and Analysis}
\label{sec:results}
In this section, we will conduct extensive experiments to evaluate the competing methods in terms of accuracy and efficiency. All experiments will be implemented in PyTorch $1.0$ on a workstation with GPU NVIDIA $2080$ti, Intel Core i$9$-$9900$KF CPU ($3.6$ GHz), $16$GB RAM, and Windows $10$ operating system.
\begin{table}[t]
	\setlength{\tabcolsep}{5pt}
	\centering
	\caption{The statistical and geometrical information related to $3$ different experimental water areas of rectangle, visually illustrated in Figs. \ref{fig:CFDexper}-\ref{fig:YREexper}. The boundary points are in the form of geographical coordinates.}
	\begin{tabular}{|c||c|c|c|c|c|c|}
		\hline
		Water Areas  & {Time Span}                                   & \tabincell{c}{Number of\\Trajectories} & \tabincell{c}{Update\\Rate}          &\tabincell{c}{Boundary\\Points} & Longitude$(^\circ)$ & Latitude$(^\circ)$ \\ \hline \hline
		\multirow{2}{*}{\tabincell{c}{Caofeidian Port}}              & \multirow{2}{*}{1/1/2018-30/1/2018} & \multirow{2}{*}{1391}        & \multirow{2}{*}{2-30s} & Left Top       & 118.291206  & 38.772082 \\ 
		&                    &           &                           & Right Bottom   & 118.558596  & 38.897793  \\ \hline
		\multirow{2}{*}{Chengshan Cape}                       & \multirow{2}{*}{1/1/2018-30/1/2018} & \multirow{2}{*}{1047}        & \multirow{2}{*}{2-30s} & Left Top       & 122.761476  & 37.523868  \\ 
		&                   &            &                           & Right Bottom   & 123.162561  & 37.715612 \\ \hline
		\multirow{2}{*}{\tabincell{c}{South Channel of\\Yangtze River Estuary}} & \multirow{2}{*}{1/8/2017-30/8/2017} & \multirow{2}{*}{1397}        & \multirow{2}{*}{2-30s} & Left Top       & 121.437999  & 31.262001  \\ 
		&                     &          &                           & Right Bottom   & 121.975002  & 31.531999 \\ \hline
	\end{tabular}\label{tab:datasets}
\end{table}
\subsection{Experimental Data Sets}
\label{sec:datasets}
To evaluate the proposed unsupervised learning method, our experiments will be performed on realistic vessel trajectories with different geometrical features. These original trajectories were collected from the terrestrial AIS base stations in $3$ different water areas, i.e., the Caofeidian Port (CFD), the Chengshan Cape (CSC), and the South Channel of Yangtze River Estuary (YRE). Table \ref{tab:datasets} detailedly depicts the statistical and geometrical information related to these experimental water areas. In our experiments, the unique nine digit identification number, i.e., Maritime Mobile Service Identity (MMSI), was adopted to classify the vessel trajectories from collected AIS data. It is easy to find that AIS records often suffer from random outliers and missing data during signal acquisition and transmission. To overcome these negative influences on similarity computation, the qualities of original trajectories have been promoted directly using the existing methods studied in previous work \citep{liang2019neural,zhang2018data}.
\subsection{Comparisons with Other Competing Methods}
In the literature \citep{su2020survey}, trajectory similarity computation methods can be mainly divided into two categories. The first category is involved to the shape-based method which considers the geometric features of trajectories. Warping-based similarity is another category of similarity computation methods, which minimizes the cumulative distance between the matched timestamped points in trajectories. To compare and evaluate the performance of similarity computation, our CAE will be compared with two representative shape- and warping-based methods, i.e., Fr{\'e}chet distance and DTW. Their definitions are briefly described as follows
\begin{itemize}
	\item \textbf{Fr{\'e}chet Distance:} The Fr{\'e}chet distance measures the spatial similarity between trajectories which takes into consideration the location and ordering of timestamped points along the trajectories. It is easier to understand through the leash-metaphor: a man traverses a path (i.e., trajectory) while walking his dog (traversing a separate path) on a leash. The distance can thus be regarded as the length of the shortest leash so that both can traverse their separate paths from start to finish \citep{buchin2016computing}.
	\item \textbf{Dynamic Time Warping (DTW):} The DTW has become one of the most popular trajectory similarity measures since 1980s. It is essentially a recursive implementation and can be easily implemented in dynamic programming. Its main objective is to search an optimal alignment between any two trajectories by warping the timestamped points iteratively \citep{de2012machine}.
\end{itemize}

To make the experiments fair, trajectory similarity computation results are yielded by these two competing methods with the optimal parameters manually selected. We refer the interested reader to see the references for more details.
\subsection{Parameter Settings}
It is well known that the choice of parameters plays an important role in our CAE-based unsupervised learning model. In this subsection, we will perform exhaustive experiments to manually determine the optimal parameters, i.e., grid size, learning rate, and dimension of feature vector. Without loss of generality, these experiments will be implemented based on the AIS-based vessel trajectories collected from YRE.
\begin{table}[t]
	\centering
	\caption{The influences of batch size and spatial interval $\Delta$ (or grid size) on CAE-based unsupervised learning. The loss function in Eq. (\ref{eq:lossfunction}) is adopted to assist in manually selecting the optimal interval $\Delta$ (or grid size) in this work.}
	\begin{tabular}{|c|c|c|c|c|}
		\hline
		\multicolumn{1}{|c|}{\multirow{2}{*}{Batch Size}} & \multicolumn{4}{c|}{Spatial Interval (Grid Size)} \\ \cline{2-5}
		\multicolumn{1}{|c|}{}                           & $0.012^{\circ}~(50 \times 37)$         & $0.009^{\circ}~(66 \times 50)$         & $0.006^{\circ}~(100 \times 75)$        & $0.003^{\circ}~(200 \times 150)$ \\ \hline
		$100$  		 & $0.0156$        	      & $0.0123$                 & $0.0154$      & $0.0124$     \\ \hline
		$200$  		 & $0.0150$        	      & $0.0103$                 & $0.0168$      & $0.0127$     \\ \hline
		$400$		 & $0.0199$        	      & $0.0139$                 & $0.0157$      & $0.0125$   \\ \hline
		$600$		 & $0.0222$        	      & $0.0128$                 & $0.0139$      & $0.0126$   \\ \hline
	\end{tabular}\label{tab:GridSize}
\end{table}
\begin{figure}[t]
	\centering\includegraphics[width=1\linewidth]{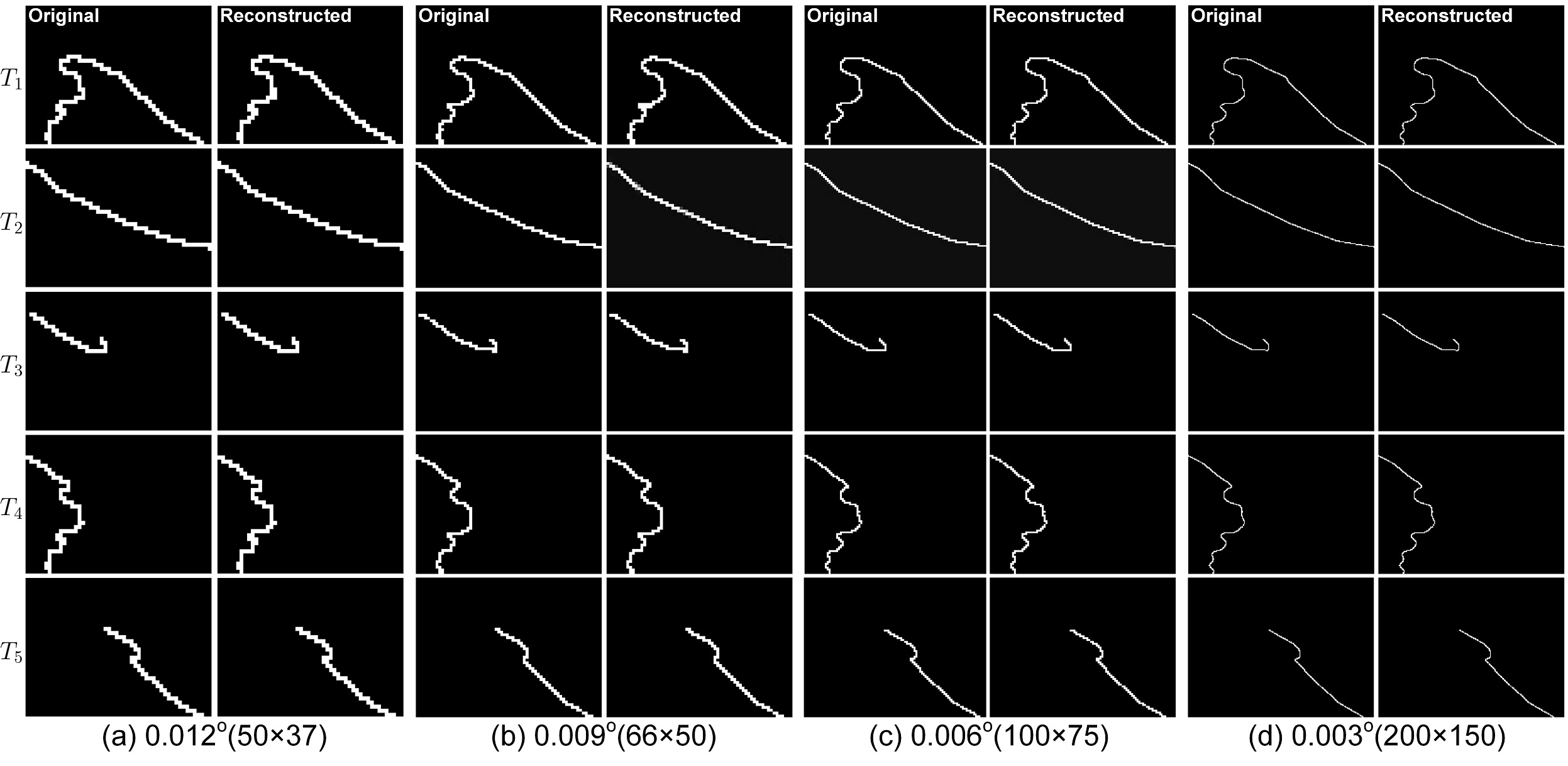}
	\caption{The visual comparisons of informative trajectory images for different spatial intervals $\Delta$ (i.e., grid sizes). From left to right: the original and reconstructed trajectory images, related to $5$ different vessel trajectories (i.e., $T_{1}, T_{2}, \cdots, T_{5}$), for (a) $0.012^{\circ}~(50 \times 37)$, (b) $0.009^{\circ}~(66 \times 50)$, (c) $0.006^{\circ}~(100 \times 75)$, and (d) $0.003^{\circ}~(200 \times 150)$, respectively.}
	\label{fig:GridSize}
\end{figure}
\subsubsection{Grid Size for Trajectory Image Generation}
The first step of our proposed framework corresponds to the image representations of vessel trajectories. In our numerical experiments, it should first divide the water areas (i.e., CFD, CSC and YRE) into a finite number of grids. The informative trajectory images will then be obtained by projecting the timestamped points in original vessel trajectories into the grids. To generate high-quality trajectory images, it is necessary to optimally select the grid size. If the grid size is large, the resolution of trajectory image will become higher, but easily leading to disconnected features which correspond to the adjacent timestamped points in raw trajectories. The computational cost in CAE training will also be extremely increased due to the high-resolution trajectory image. In contrast, the small grid size will yield low-resolution trajectory image, easily resulting in loss of small-scale geometrical features in raw trajectories. The stability and accuracy of similarity computation will be degraded accordingly. To enhance the robustness of similarity measure, the cubic spline interpolation method has been introduced to rearrange the raw AIS data with the time interval being $5$s between any two adjacent points \footnote{Experimental results illustrate that the cubic spline interpolation is able to guarantee accurate and robust computation of similarities between different vessel trajectories in different water areas.}.

As shown in Fig. \ref{fig:TrajImage}, the grid size is essentially related to the spatial interval when dividing the water areas into a finite number of grids. The influences of batch size and spatial interval $\Delta$ (or grid size) on CAE-based unsupervised learning are summarized in Table \ref{tab:GridSize}. It can be found that our CAE model can achieve the lowest loss value if batch size = $200$ and $\Delta = 0.009^{\circ}$. We further visually compare the informative trajectory images for different spatial intervals (i.e., grid sizes) with the fixed batch size of $200$. The original trajectory images and their reconstructed versions\footnote{Note that original trajectory images are generated by directly projecting the original vessel trajectories with time interval of $5$s into the discrete water areas. In contrast, reconstructed trajectory images are obtained through the decoder network introduced in Section \ref{sec:CAEmodel}.} are illustrated in Fig. \ref{fig:GridSize}. The original and reconstructed images have similar geometrical structures for different spatial intervals or grid sizes. It seems that our CAE model performs well in low-dimensional representations of original vessel trajectories and high-quality reconstruction of informative trajectory images from the learned low-dimensional features. Without loss of generality, we propose to directly select the batch size = $200$ and $\Delta = 0.009^{\circ}$ (i.e., grid size of $66 \times 50$) in our experiments on similarity computation. Experimental results have demonstrated the effectiveness and robustness of these manually-selected parameters.
\begin{figure}[t]
	\centering\includegraphics[width=0.7\linewidth]{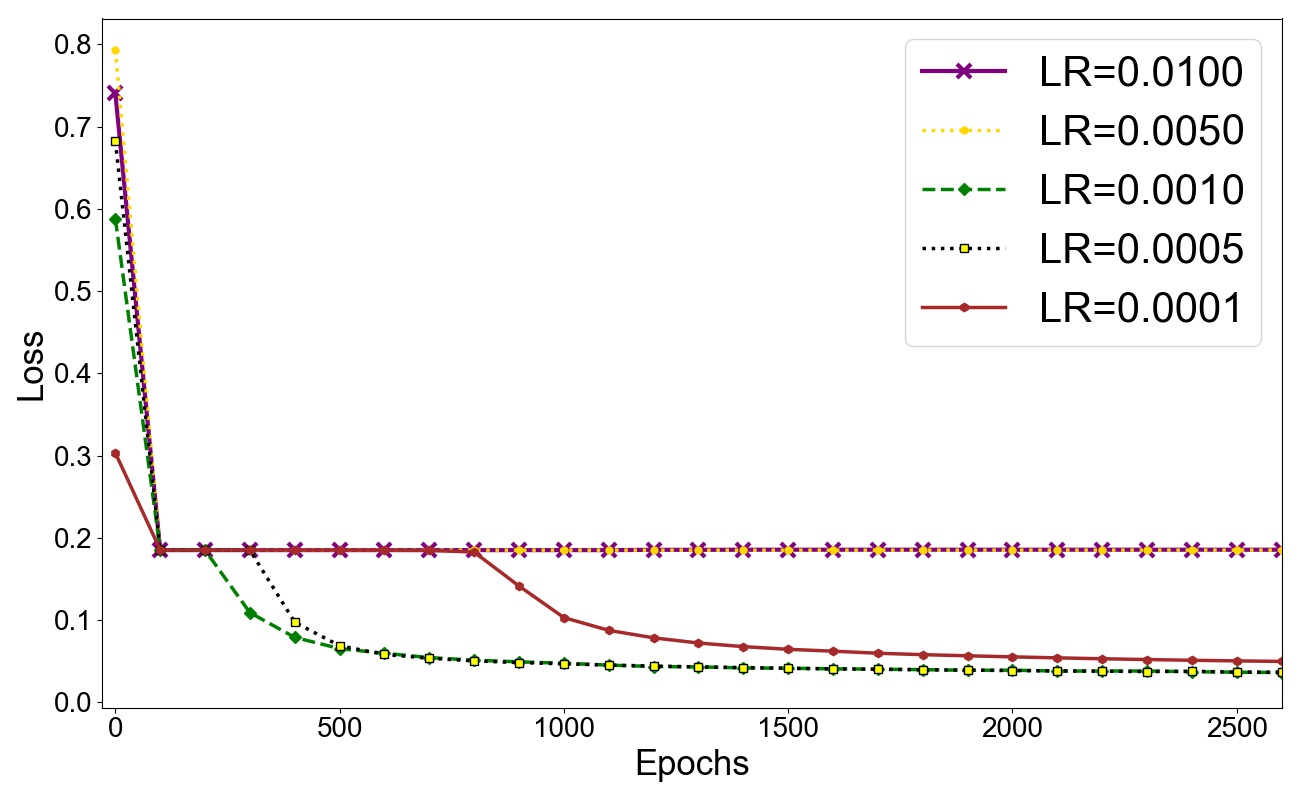}
	\caption{The influences of learning rates (e.g., $\mathrm{LR} = 0.0005$, $0.0010$, $0.0050$, $0.0060$ and $0.0070$) on loss during CAE unsupervised training.}
	\label{fig:LearningRate}
\end{figure}
\subsubsection{Learning Rate for CAE Unsupervised Training}
It is well known that tuning the learning rate (LR) or selecting an optimal LR in CAE is still a challenging problem. Depending upon the value of LR, the accuracy of similarity computation can vary tremendously. Generally, a large LR allows our unsupervised learning model to accelerate training process considerably, but easily causes instability. In contrast, a smaller LR can allow our model to learn an optimal set of weights but at the cost of significantly long training time. Many efforts have been focused on adaptive LR \citep{liang2019barzilai}, it is still difficult to automatically determine the optimal LR in maritime applications. In this work, we propose to manually try some LR values and select the optimal one. Extensive experiments have been performed to investigate different LR values on CAE unsupervised training. As shown in Fig. \ref{fig:LearningRate}, if the LR values are less than or equal to $0.001$, it becomes easier to guarantee the robust convergence of our CAE training. Furthermore, we find that $0.001$ is better to train our CAE-based unsupervised learning model. The vessel trajectory similarity computation results under this parameter setting are consistently promising in different water areas.
\begin{figure}[t]
	\centering\includegraphics[width=1\linewidth]{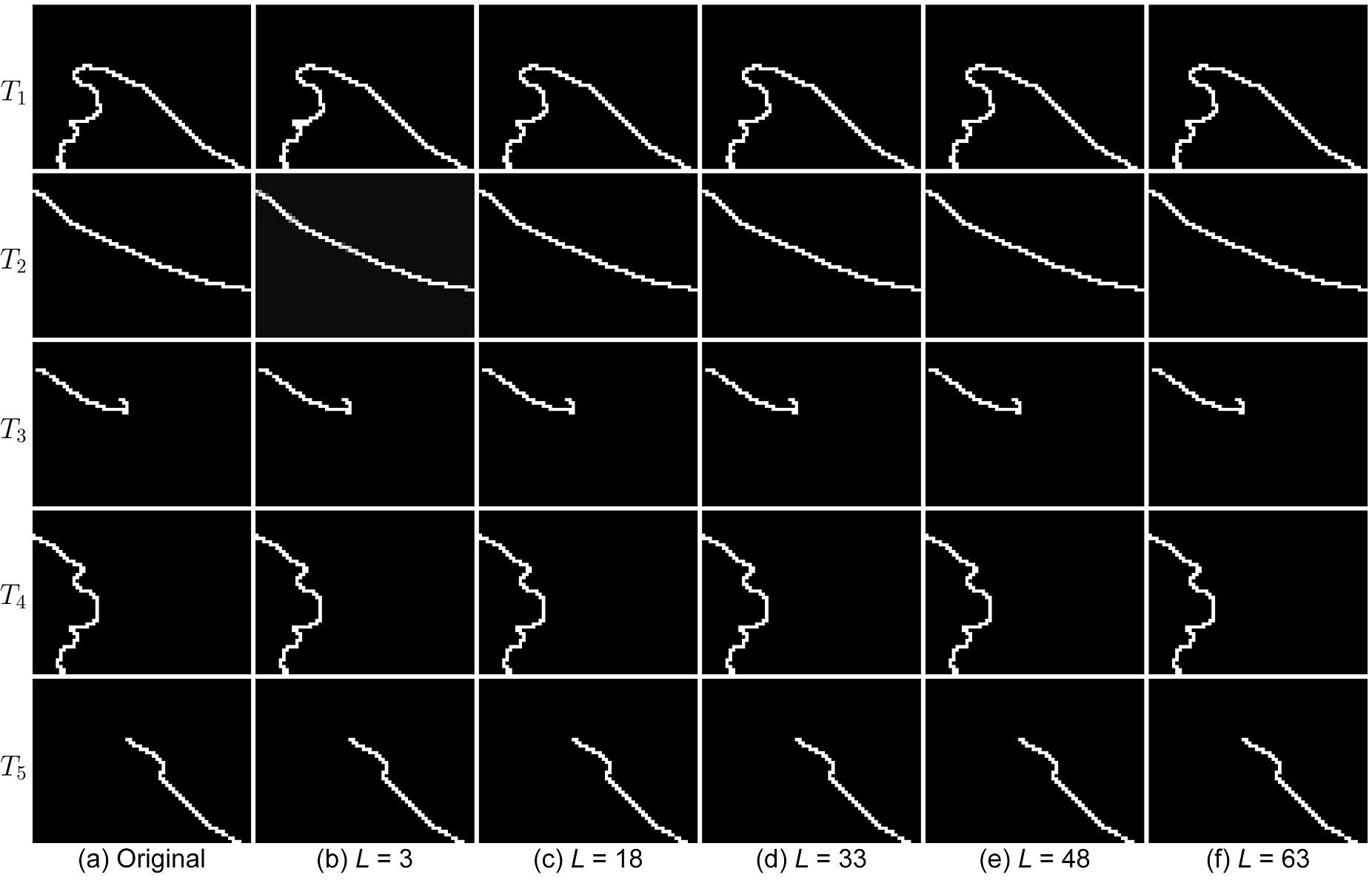}
	\caption{The influences of different dimensions (e.g., $L = 3$, $18$, $33$, $48$ and $63$) on reconstruction of informative trajectory images from the low-dimensional feature vectors. From left to right: (a) original informative trajectory images, and their reconstructed versions from (b) $3$-, (c) $18$-, (d) $33$-, (e) $48$, and (f) $63$-dimensional features vectors, respectively.}
	\label{fig:FeatureLength}
\end{figure}
\subsubsection{Dimension of Feature Vectors for Low-Dimensional Representation}
\label{sec:DimFeature}
The optimal selection of the dimension of feature vectors also plays an important role in CAE-based trajectory similarity computation. Theoretically, the larger dimension of feature vectors brings more accurate measure of trajectory similarity, but leading to improved computational cost. In contrast, the lower dimension can significantly reduce computational time, but may be resulting in trajectory similarity degradation. Therefore, it is necessary to appropriately determine the dimension of feature vectors in our CAE-based unsupervised learning method. In this work, we mainly investigate the influences of different dimensions (e.g., $L = 3$, $18$, $33$, $48$ and $63$) on reconstruction of informative trajectory images from the low-dimensional feature vectors. The experimental results in Fig. \ref{fig:FeatureLength} show that the mentioned dimensions are capable of accurately reconstructing the vessel trajectories. It is thus sufficient to represent the vessel trajectories (or informative trajectory images) using our CAE model with only $L = 3$. In the following numerical experiments, $L = 3$ is directly selected as the dimension of feature vectors to shorten computational time while guaranteeing high-quality similarity computation.
\begin{table}[t]
	\centering
	\caption{Comparisons of computational costs for different methods (i.e., DTW, Fr{\'e}chet, and CAE) under different numbers of vessel trajectories (i.e., $100$, $200$, $300$, $400$, and $500$). $N$ denotes the number of vessel trajectories randomly selected in our experiments. $R_{\mathrm{D/C}}$ and $R_{\mathrm{F/C}}$, respectively, represent the ratios of the exection time of DTW and Fr{\'e}chet to that of CAE for the same trajectory similarity measure task. To reduce the randomness, each similarity measure method ran $100$ times to obtain the average results (i.e., mean $\pm$ std.).}
	\begin{tabular}{|c||c|c|c||c|c|}
		\hline
		$N$ & DTW & Fr{\'e}chet & CAE & $R_{\mathrm{D/C}}$ & $R_{\mathrm{F/C}}$ \\ \hline
		$100$ & $27.162 \pm 2.635$ & $21.483 \pm 2.074$  &  $0.58 \pm 0.02$ & $46.822 \pm 4.86$ & $37.028 \pm 3.78$  \\ \hline
		$200$ & $109.13 \pm 7.409$ & $86.079 \pm 5.907$  &  $1.18 \pm 0.03$ & $92.345 \pm 6.91$ & $72.816 \pm 5.16$  \\ \hline
		$300$ & $244.57 \pm 13.13$ & $193.09 \pm 10.22$  &  $1.93 \pm 0.08$ & $126.81 \pm 8.75$ & $100.11 \pm 6.69$ \\ \hline
		$400$ & $427.32 \pm 17.38$ & $344.24 \pm 15.72$  &  $2.76 \pm 0.04$ & $155.07 \pm 6.36$ & $124.93 \pm 5.94$ \\ \hline
		$500$ & $672.75 \pm 22.06$ & $534.99 \pm 18.15$  &  $3.72 \pm 0.04$ & $181.00 \pm 6.38$ & $143.92 \pm 4.87$ \\ \hline
	\end{tabular}\label{tab:RunningTime}
\end{table}
\subsection{Experiments on Computational Efficiency}
Both DTW and Fr{\'e}chet are essentially limited by the bottleneck in high computational cost. Many extensions \citep{begum2015, gudmundsson2019fast} have been proposed to accelerate similarity computation. However, the computational process is still extremely time consuming, especially for massive vessel trajectories in maritime data mining. To evaluate the computational efficiency, we tend to calculate the execution times of similarity computation on different numbers of vessel trajectories. In particular, to achieve impartial experimental results, we randomly select $100$-$500$ vessel trajectories with different timestamped points. Each similarity computation experiment runs $100$ times for each experimental scenario. The computational costs for DTW, Fr{\'e}chet, and our CAE have been detailedly summarized in Table \ref{tab:RunningTime}. It can be found that the running time for both DTW and Fr{\'e}chet increases significantly with the number of vessel trajectories becoming larger. In contrast, our CAE can tremendously reduce the computational time even in the case of massive vessel trajectories. As the number of trajectories becomes larger, our superior computational performance will be more obvious. The high computational costs will hinder the applications of both DTW and Fr{\'e}chet in maritime practice. Benefiting from the low-dimensional feature vectors, our CAE is capable of efficiently computing the similarities between different vessel trajectories. This advantage will become more significant in the case of massive vessel trajectories, which is more common in maritime data mining. The efficient similarity computation is of fundamental importance to the fast detection of deviating and abnormal vessel behaviors in maritime surveillance.
\subsection{Experiments on Computational Accuracy}
\subsubsection{Assessment Criterion}
To the best of our knowledge, there is no reliable assessment criterion to quantitatively evaluate the accuracy of trajectory similarity computation thus far. To handle this shortcoming, we propose to select the trajectory clustering results to indirectly evaluate the similarity. Note that the clustering results are highly dependent upon the similarity computation. If we can accurately measure the trajectory similarities, the high-quality clustering results can be obtained accordingly. The purpose of clustering is to divide the given objects into a number of groups such that similar objects belong to the same group whereas the dissimilar objects are in different groups. In this work, the vessel trajectories in the same group (i.e., cluster) will have high similarities.

Motivated by previous study \citep{besse2016review}, we propose to jointly adopt the Between-Like (BC) and Within-Like (WC) criteria to assess trajectory similarity. For a predefined number of clusters, if we can generate accurate similarity measure, the WC criterion should be as small as possible, and the BC criterion should be as large as possible. In particular, the definitions of BC and WC are, respectively, given by
\begin{equation}\label{eq:BC}
    \mathrm{BC} = \sum_{z = 1}^{Z} \mathcal{D}_{\mathcal{E}} \left( \mathcal{H} \left( \mathcal{X}^{\mathrm{mean}}_{\mathcal{T}} \right), \mathcal{H} \left( \mathcal{X}^{\mathrm{mean}}_{C_{z}} \right) \right),
\end{equation}
\begin{equation}\label{eq:WC}
    \mathrm{WC} = \sum_{z = 1}^{Z} \frac{1}{|C_{z}|} \sum_{T_{i} \in C_{z}} \mathcal{D}_{\mathcal{E}} \left( \mathcal{H} \left( \mathcal{X}^{\mathrm{mean}}_{C_{z}} \right), \mathcal{H} \left( \mathcal{X}_{i} \right) \right),
\end{equation}
with informative trajectory images $\mathcal{X}_{i}$, $\mathcal{X}^{\mathrm{mean}}_{\mathcal{T}}$, and $\mathcal{X}^{\mathrm{mean}}_{C_{z}}$ being generated from $T_{i}$, $T^{\mathrm{mean}}_{\mathcal{T}}$, and $T^{\mathrm{mean}}_{C_{z}}$, respectively. Here, $C_z$ is the $z$-th cluster, $Z$ is the total number of clusters, $T^{\mathrm{mean}}_{ \circ }$ represents the mean of vessel trajectories, and $\mathcal{T}$ denotes the set of vessel trajectories considered in our experiments.

It is mathematically intractable to calculate the mean of vessel trajectories collected from $\mathcal{T}$ or $C_{z}$ with $z \in \{1, 2, \cdots ,Z\}$. In this work, we tend to select an exemplar $T^{\mathrm{ex}}_{ \circ }$ to approximate $T^{\mathrm{mean}}_{ \circ }$, i.e., $T^{\mathrm{mean}}_{\mathcal{T}} \leftarrow T^{\mathrm{ex}}_{\mathcal{T}}$ and $T^{\mathrm{mean}}_{C_{z}} \leftarrow T^{\mathrm{ex}}_{C_{z}}$. Taking $\mathcal{T}$ as an example, we can define $T^{\mathrm{ex}}_{ \mathcal{T} }$ as follows
\begin{equation}\label{eq:Exemplar}
    T^{\mathrm{ex}}_{ \mathcal{T} } = \min_{T_{i}} \left\{ \sum_{j = 1, j \neq i}^{M} \mathcal{D}_{\mathcal{E}} \left( \mathcal{H} \left( \mathcal{X}_{i} \right), \mathcal{H} \left( \mathcal{X}_{j} \right) \right) \right\},
\end{equation}
for $i \in [1, 2, \cdots, M ]$ with $M$ being the total number of vessel trajectories in data set $\mathcal{T}$.

If the criteria BC and WC are obtained, we then propose the following assessment criterion (AC) to quantitatively evaluate the accuracy of trajectory similarity measure, i.e.,
\begin{equation}\label{eq:AC}
    \mathrm{AC} = \frac{ \mathrm{WC} }{ \mathrm{BC} + \mathrm{WC} } \in \left( 0, 1 \right).
\end{equation}

This criterion is essentially related to the trajectory similarity measure. If we can obtain accurate similarity measure, the high-quality clustering results can be accordingly generated leading to lower value of AC. In contrast, higher value of AC indicates the inaccurate computation of similarities between different vessel trajectories. From a theoretical point of view, as the number of clusters increases, the value of AC will be decreased correspondingly. However, the large number could bring the negative effects on modeling ship traffic behavior, resulting in unreliable detection of deviating or abnormal vessel behaviors. The obtained AC could enable the proper selection of the number of clusters. Once the value of AC meets the accuracy requirement, it is better to choose the number of clusters as small as possible.
\begin{figure}[t]
	\centering\includegraphics[width=1\linewidth]{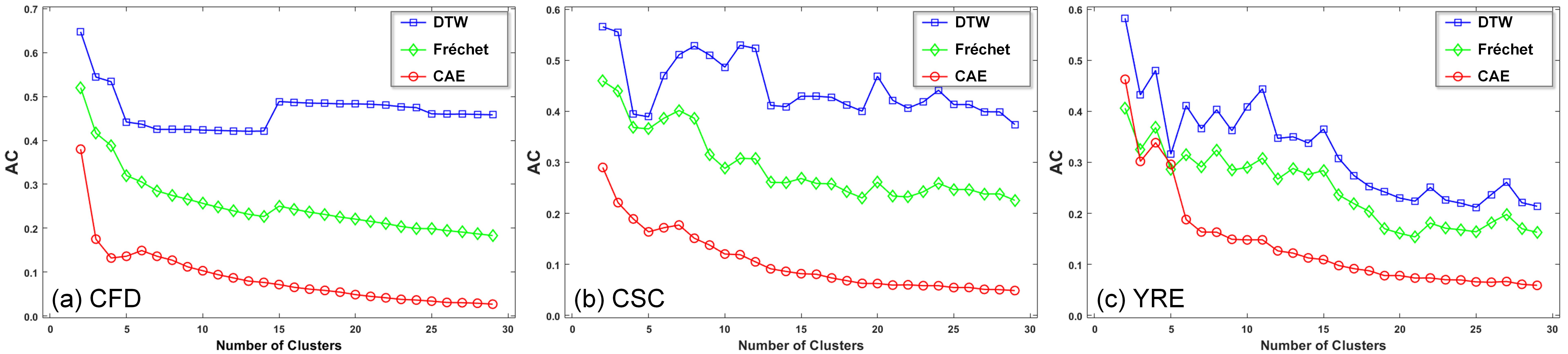}
	\caption{The Results of $AC$ Criterion Obtained from CAE, Fr{\'e}chet, DTW with the Number of Clusters. From left to right: The CFD, CST and EYR datasets were used to evaluate the quality of trajectory similarity computation, respectively.}\label{fig11}
\end{figure}
\begin{figure}[t]
	\centering\includegraphics[width=1\linewidth]{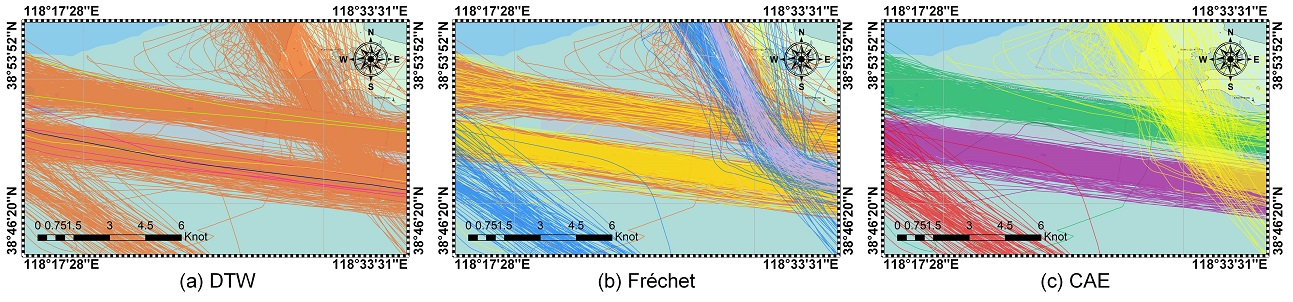}
	\caption{Visual comparisons of vessel trajectory clustering results for CFD data set. From left to right: clustering results based on trajectory similarities generated by (a) DTW, (b) Fr{\'e}chet, and (c) CAE, respectively.}\label{fig:CFDexper}
\end{figure}
\begin{figure}[t]
	\centering\includegraphics[width=1\linewidth]{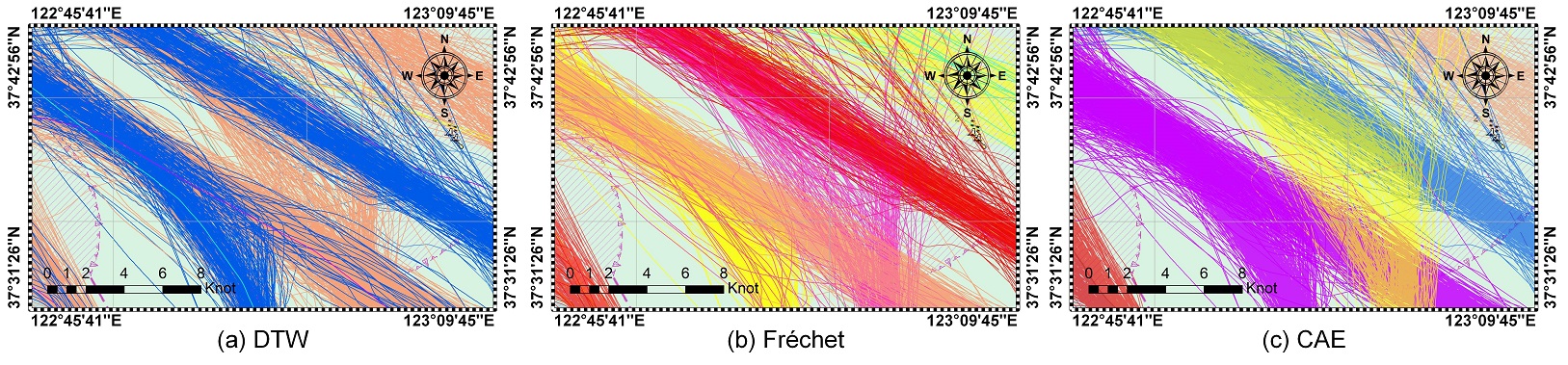}
	\caption{Visual comparisons of vessel trajectory clustering results for CSC data set. From left to right: clustering results based on trajectory similarities generated by (a) DTW, (b) Fr{\'e}chet, and (c) CAE, respectively.}\label{fig:CSCEexper}
\end{figure}
\begin{figure}[t]
	\centering\includegraphics[width=1.0\linewidth]{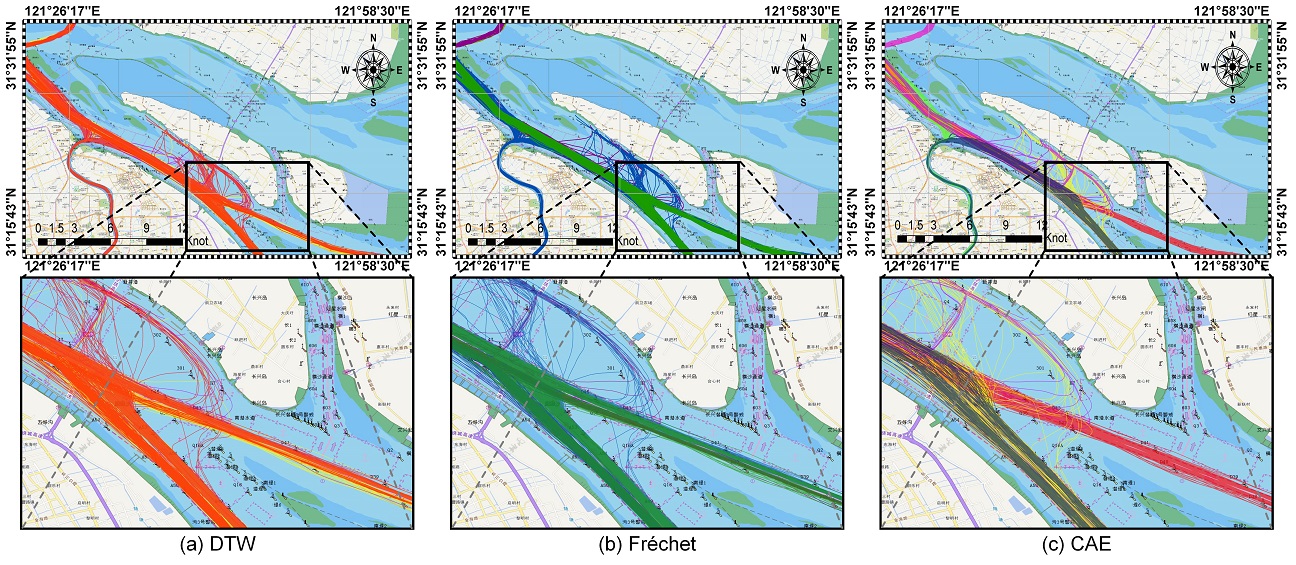}
	\caption{Visual comparisons of vessel trajectory clustering results for YRE data set. From left to right: clustering results based on trajectory similarities generated by (a) DTW, (b) Fr{\'e}chet, and (c) CAE, respectively.}\label{fig:YREexper}
\end{figure}
\begin{figure}[t]
	\centering\includegraphics[width=1.0\linewidth]{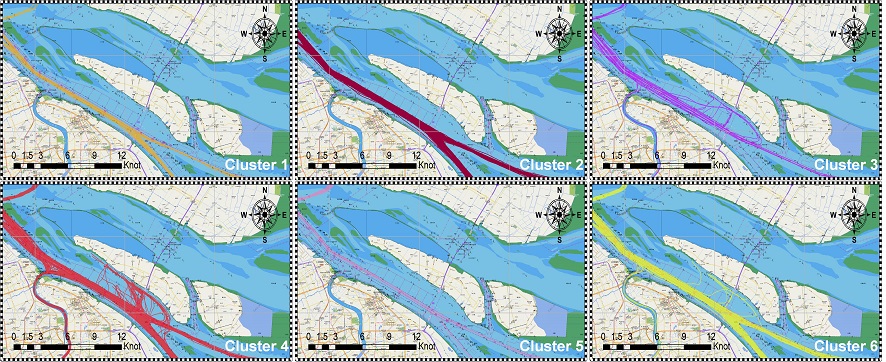}
	\caption{Visualizations of $6$ different clusters of vessel trajectories, obtained through the DTW-based trajectory similarities.}\label{fig:Clusters01DTW}
\end{figure}
\subsubsection{Comparisons of Similarity Computations}
Clustering experiments are implemented on $3$ different water areas (i.e., CFD, CSC and YRE) to assess the trajectory similarity computation results. As discussed beforehand, according to the experiment results, we manually select the optimal parameters (i.e., batch size = $200$, grid size = $66 \times 50$, LR = $0.006$, and epoch = $3000$) for CAE training. The dimension of feature vectors, learned from the informative trajectory images, is set to be $3$ in trajectory similarity computation. In particular, the similarities between any two low-dimensional feature vectors are computed by Euclidean distance operator. The final clustering results are obtained using HCA according to the similarities computed by DTW, Fr{\'e}chet, and CAE, respectively.

In order to select a reasonable number of clusters and evaluate the effect of clustering, the number of clusters are ranges from 2 to 25 to get different experiment results. Those results would be evaluated by AC criterion, creating the line diagram shown as Fig. \ref{fig11}. As observed from Fig. \ref{fig11} (a), when the number of clusters is around $4$, our CAE-based learning model could generate the satisfactory AC criterion which tends to be changed slightly as the number of clusters increases in CFD data set. In theory, the larger number of clusters is able to divide original vessel trajectories into more partitions. It is benefit for identifying more vessel moving patterns in maritime supervision. However, the larger number could cause one cluster to be separated into different clusters. For the sake of simplicity, we tend to directly set the number of clusters to be $4$ to cluster vessel trajectories. Analogous to the selection process in CFD data set, we empirically determine the numbers of clusters as $5$ and $6$ for vessel trajectory clustering in CSC and YRE data sets, respectively.
\begin{figure}[t]
	\centering\includegraphics[width=1.0\linewidth]{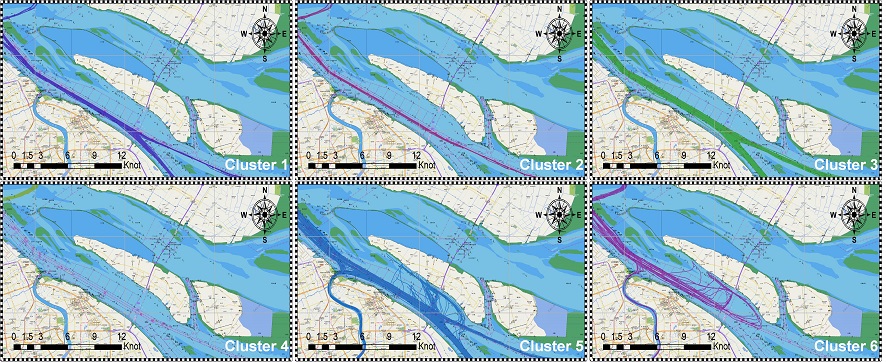}
	\caption{Visualizations of $6$ different clusters of vessel trajectories, obtained through the Fr{\'e}chet-based trajectory similarities.}\label{fig:Clusters02Frechet}
\end{figure}
\begin{figure}[t]
	\centering\includegraphics[width=1.0\linewidth]{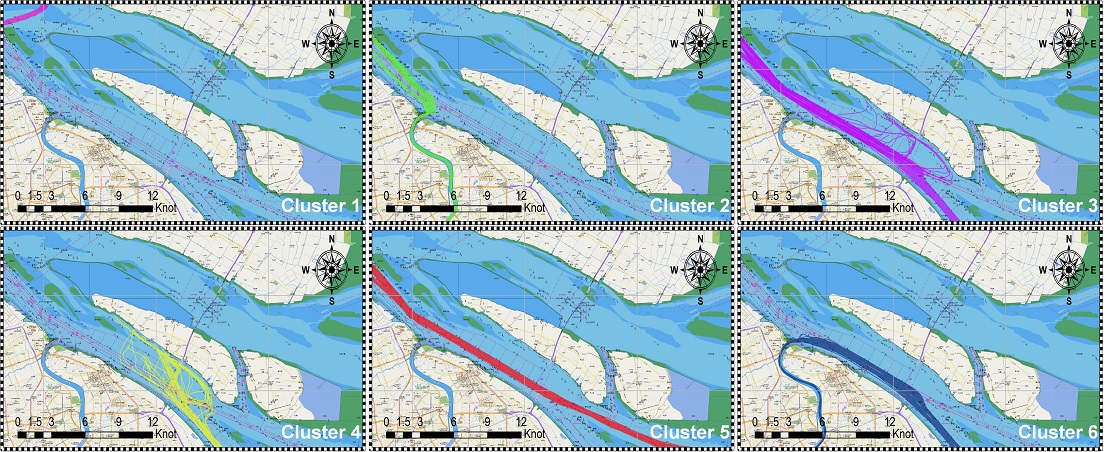}
	\caption{Visualizations of $6$ different clusters of vessel trajectories, obtained through the CAE-based trajectory similarities.}\label{fig:Clusters03CAE}
\end{figure}

To further illustrate the effectiveness of our proposed approach, we tend to show the visualization of vessel trajectory clustering for comparison. The visual results of clustered vessel trajectories are presented in Figs. \ref{fig:CFDexper}-\ref{fig:YREexper}, which verify the superiorities of our CAE network in enhancing trajectory clustering. Both CFD and CSC data sets are selected because of the trajectory clusters easy to generate. In contrast, it is difficult to cluster the vessel trajectories for the complex maritime environments in YRE data set. It is clear to observe that trajectory clustering results, generated by DTW- and Fr{\'e}chet-based HCA methods, suffer from obvious clustering mistakes in Figs. \ref{fig:CFDexper} and \ref{fig:CSCEexper}. Owing to the superior feature learning, our CAE-based method performs well in generating high-quality trajectory clustering performance. The advantage of our method is further confirmed by the more challenging clustering task in Fig. \ref{fig:YREexper}. Due to the complex navigation conditions in crossing waters, more erroneous clustering is significantly generated by both DTW- and Fr{\'e}chet-based clustering methods. In contrast, the proposed method has the capacity of explicitly clustering the complicated vessel trajectories. It is able to accurately distinguish the traffic patterns of moving vessels in crossing waters and different water channels. The local magnification views shown in Fig. \ref{fig:YREexper} illustrate that our method generates a more natural-looking clustering results compared with other competing methods. To further demonstrate our superior performance, the obtained $6$ different clusters of vessel trajectories in Fig. \ref{fig:YREexper} are respectively visualized in Figs. \ref{fig:Clusters01DTW}-\ref{fig:Clusters03CAE}. It can be found that our proposed method can significantly improve the trajectory clustering performance. Both DTW- and Fr{\'e}chet-based HCA methods easily cause inaccurate clusters due to the unreliable trajectory similarities between different trajectories. Benefiting from the accurate similarity measure by CAE, our method is capable of guaranteeing robust and reliable trajectory clustering results. The high-quality trajectory clustering results obtained play important roles in maritime applications, e.g., route planning, maritime supervision, and abnormal behavior detection, etc.
\section{Conclusions and Future Perspectives}
\label{sec:conclusion}
We have proposed a CAE-based unsupervised learning method for computing similarities between different vessel trajectories. First, the original vessel trajectories were remapped into two-dimensional matrices (i.e., informative trajectory images). Next, low-dimensional representations of the informative trajectory images were then obtained using the CAE network. The similarities between different vessel trajectories were then equivalent to calculating the distances between the corresponding low-dimensional feature vectors. Experiments demonstrated that our proposed method obviously outperformed other competing methods in terms of both efficiency and effectiveness. The superior trajectory clustering results could also be accordingly guaranteed for different data sets.

It should be noted that there is still lack of standard benchmarks related to vessel trajectories, directly adopted to quantitatively evaluate the accuracy of similarity computation. Although the indirect evaluation manner, i.e., trajectory clustering performance, could be considered a feasible method in this work. Experiments on standard benchmarks could theoretically bring more reliable and accurate results for different vessel trajectory similarity computation methods. There is thus a significant potential to design standard benchmarks for vessel trajectories (with labels). To further enhance the accuracy of trajectory similarity computation, it not only considers the spatial information of timestamped points, but also the course and speed of vessels. These dynamic characteristics are essentially related to the navigation behaviors of vessels. The combination of position, course and speed of vessels could hence promote the vessel trajectory similarity computation in maritime applications.
\section*{Acknowledgments}
The work described in this paper was supported by grants from the State Key Laboratory of Resources and Environmental Information System, and the National Key R\&D Program of China (No.: 2018YFC0309602).
\bibliographystyle{cas-model2-names}

\bibliography{cas-refs}


%
%


\end{document}